\documentclass[]{uiuc_paper}
\pdfoutput=1
\usepackage[toc,page,header]{appendix}
\usepackage{cleveref}
\usepackage{subcaption}
\usepackage{booktabs}
\usepackage{tabularx}

\usepackage{latexsym}
\usepackage{url}
\usepackage{amssymb}
\usepackage{amsmath}
\usepackage{amsfonts}
\usepackage[utf8]{inputenc}
\usepackage{booktabs}
\usepackage{pifont}
\usepackage{multirow}
\usepackage{makecell}
\usepackage{paralist}
\usepackage{xspace}
\usepackage{adjustbox}

\usepackage{colortbl}
\usepackage{algorithm}
\usepackage{algpseudocode}
\usepackage{capt-of}
\crefname{section}{Sec.}{Secs.}\Crefname{section}{Sec.}{Secs.}
\crefname{subsection}{Sec.}{Secs.}\Crefname{subsection}{Sec.}{Secs.}
\crefname{subsubsection}{Sec.}{Secs.}\Crefname{subsubsection}{Sec.}{Secs.}
\crefname{appendix}{App.}{Apps.}\Crefname{appendix}{App.}{Apps.}
\crefname{subappendix}{App.}{Apps.}\Crefname{subappendix}{App.}{Apps.}
\crefname{subsubappendix}{App.}{Apps.}\Crefname{subsubappendix}{App.}{Apps.}
\crefname{figure}{Fig.}{Figs.}\Crefname{figure}{Fig.}{Figs.}
\crefname{table}{Tab.}{Tabs.}\Crefname{table}{Tab.}{Tabs.}
\crefname{equation}{Eq.}{Eqs.}\Crefname{equation}{Eq.}{Eqs.}
\crefname{algorithm}{Alg.}{Algs.}\Crefname{algorithm}{Alg.}{Algs.}
\creflabelformat{equation}{#2#1#3}
\definecolor{oursbg}{HTML}{E4F3F3}
\definecolor{gain}{HTML}{1F7F7E}
\definecolor{loss}{HTML}{B0596B}
\definecolor{cited}{HTML}{636E7B}
\newcommand{\AbsResults}{With Qwen2.5-1.5B it lifts GRPO by over twenty points on ALFWorld and over ten on WebShop, overtakes GiGPO on ALFWorld and learns in far fewer iterations; it also improves GiGPO and GraphGPO on both benchmarks and reaches the highest success of our comparison, ahead of critic-based PPO and of state-of-the-art GiGPO, HGPO and GraphGPO.}
\newcommand{\AheadWsClause}{ and on WebShop}
\newcommand{\AlfGjPairSets}{sets B and C}
\newcommand{\AlfIterWords}{in little more than half the iterations that GiGPO needs}
\newcommand{\AlfJevGigpoTypesAheadOod}{all six}
\newcommand{\AlfJevGrpoPairId}{23.7}
\newcommand{\AlfJevGrpoPairIdHi}{30.3}
\newcommand{\AlfJevGrpoPairIdLo}{17.3}
\newcommand{\AlfJevGrpoPairOod}{29.2}
\newcommand{\AlfJevGrpoPairOodHi}{35.1}
\newcommand{\AlfJevGrpoPairOodLo}{23.4}
\newcommand{\AlfJevGrpoTypesAhead}{every}
\newcommand{\AlfJevOverGrpoId}{23.7}
\newcommand{\AlfJevOverGrpoOod}{29.2}
\newcommand{\AlfLiftSeenWords}{more than twenty}
\newcommand{\AlfLiftUnseenWords}{nearly thirty}
\newcommand{\AlfValGigpoIterEighty}{120}
\newcommand{\AlfValJevIterEighty}{70}
\newcommand{\BestRowsClause}{ In the All, Unseen, Score and Succ.\ columns the best row is a \ours{} row, and on unseen rooms all three \ours{} rows are above the best baseline.}
\newcommand{\ConclResults}{It lifts GRPO by a wide margin on ALFWorld and on WebShop and overtakes GiGPO on ALFWorld; it also improves GiGPO and GraphGPO on both benchmarks and reaches the highest success of our comparison, ahead of critic-based PPO and of the state-of-the-art GiGPO, HGPO and GraphGPO.}
\newcommand{\ContribResults}{ On ALFWorld \ours{} lifts GRPO by more than twenty points and past GiGPO; on WebShop it lifts GRPO by more than ten points and learns faster. It also improves GiGPO and GraphGPO on both benchmarks, and the best rows of \cref{tab:main} are \ours{} rows, ahead of critic-based PPO and of GiGPO, HGPO~\cite{he2026hgpo} and GraphGPO}
\newcommand{\GpuHours}{1{,}120}
\newcommand{\GpuHoursAll}{1{,}970}
\newcommand{\PlugSentences}{ It improves the GiGPO and the GraphGPO it is added to on both benchmarks: every gain in the All, Unseen, Score and Succ.\ columns of \cref{tab:main} is positive. \ours{}$_{\text{GraphGPO}}$ has the highest seen-room success of the table and the best WebShop row.}
\newcommand{\PpoSentenceB}{ GPT-4o and Gemini-2.5-Pro, prompted without training, and PPO, which fits a critic to obtain a step-level signal, are far below every \ours{} row.}
\newcommand{\TabGreySources}{GPT-4o, Gemini-2.5-Pro, PPO, RLOO~\cite{rloo}: Feng et al.~\cite{feng2025gigpo}; HGPO (two history steps), $^{a}$: He et al.~\cite{he2026hgpo}}
\newcommand{\WsAppCkptClause}{ with GiGPO at every checkpoint, with GraphGPO at five of the six.}
\newcommand{\WsEndSentence}{ \ours{}$_{\text{GRPO}}$ ends above GiGPO and more than ten points above the GRPO it is added to.}
\newcommand{\WsHalfWayClause}{ halfway through training it is already above the success that GRPO reaches at the end.}
\newcommand{\WsJevOverGrpo}{13.1}
\newcommand{\WsLiftWords}{more than ten}
\newcommand{\WsOthersSentence}{. With GiGPO and GraphGPO the credit leads early as well (\cref{app:ws}).}
\newcommand{\WsSplitSentence}{ The gain is largest in fashion, the biggest category, and where the goal demands the most: with three options to select it is more than twenty points (\cref{fig:ws}b,~c)}

\newcommand{\ours}{\textup{\textsc{DM-PRM}}}

\hypersetup{
  bookmarksnumbered=true,
  pdftitle={Credit Assignment Has Been Addressed via Decision Models},
  pdfauthor={Jiajun Fan, Qinglong Zheng, Ziwen Wang, Bangji Yang, Neil He, Ruihan Guo, Chaoran Cheng, Ge Liu}}

\title{Credit Assignment Has Been Addressed via Decision Models}

\author[1]{Jiajun Fan}
\author[1]{Qinglong Zheng}
\author[1,2]{Ziwen Wang}
\author[1]{Bangji Yang}
\author[1,3]{Neil He}
\author[1]{Ruihan Guo}
\author[1]{Chaoran Cheng}
\author[1]{Ge Liu}

\affiliation[1]{\affillogo{\includegraphics{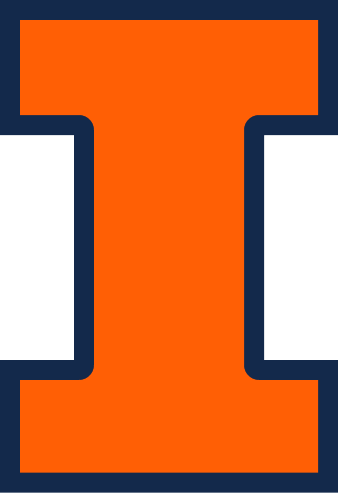}}{UIUC}}
\affiliation[2]{\affillogo{\includegraphics{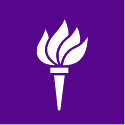}}{NYU}}
\affiliation[3]{\affillogo{\includegraphics{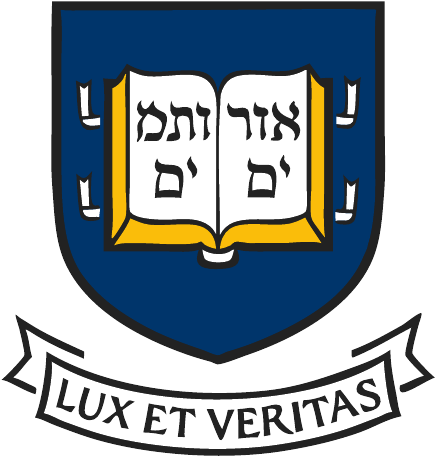}}{Yale}}

\abstract{Reinforcement learning (RL) from verifiable outcomes gives a language agent one reward per episode: every step is
credited alike, and nothing about the task is learned when all rollouts fail. The last mile, step-level credit, usually requires
step labels, a learned critic, costly LLM judges or hand-written rules. We introduce \mbox{\ours{}}, which turns
a \emph{general-purpose decision model}~-- one answering yes/no questions with probabilities~-- into
a \emph{general-purpose process reward model} (PRM), zero-shot: no step labels, no fine-tuning, no critic. \ours{}
asks one question per step and adds the standardised answers to the advantage. It plugs into any policy-gradient
method; one model, with one question text per environment, serves household tasks, web shopping and multiple RL
methods. \AbsResults{} Its scores locate useful steps in failed rollouts, and trained agents drop idle actions and
explore instead of circling. \ours{} shows that step credit need not be learned: it can be asked for.
}

\date{\today}

\begin{document}

\maketitle

\begin{figure}[t]
\centering
\includegraphics[width=\textwidth]{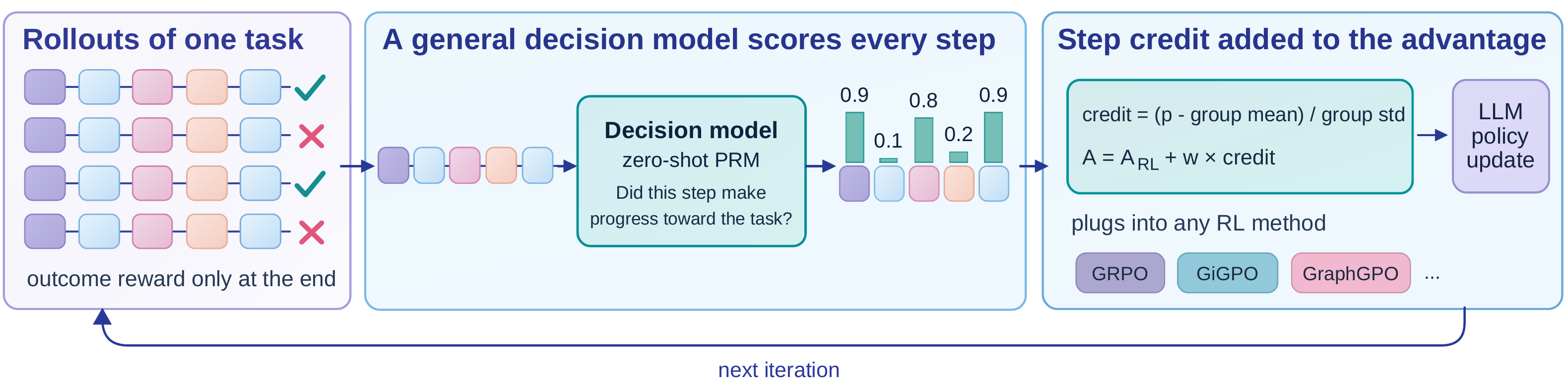}
\caption{\ours{}. The rollouts of a task receive an outcome only at the end. A general-purpose decision model, used as a
process reward model without step labels, returns for every step the probability that it made progress. The scores
are standardised against comparable steps of the same task and added, as a step credit, to the advantage of the
RL method in use; training is otherwise unchanged (pseudocode: \cref{app:algo}).}
\label{fig:framework}
\end{figure}

\section{Introduction}

\begin{flushright}
\begin{minipage}{0.64\linewidth}\raggedleft\small
\emph{``\ldots each ultimate success (or failure) is associated with a vast number of internal decisions. If the run is
successful, how can we assign credit for the success among the multitude of decisions?''}\\[2pt]
--- Marvin Minsky, 1961
\end{minipage}
\vspace{4pt}
\end{flushright}

Reinforcement learning (RL) from verifiable outcomes is now a standard way to post-train language agents, and credit
assignment, the last mile from the outcome of an episode to its individual steps, has long been its weak point. The
verifier speaks once per rollout: a fifty-step episode receives one number, and nothing says which steps earned it.
GRPO~\cite{shao2024deepseekmath} credits every step alike and has to learn from differences between many rollouts;
when all rollouts of a task fail (9 of 16 task groups in the first ALFWorld iteration, 14 of 16 on WebShop), the
group carries no signal about the task.

Existing remedies have a price. PPO~\cite{schulman2017ppo} fits a critic alongside the policy, a second network
with its own compute and instability. GiGPO~\cite{feng2025gigpo}, HGPO~\cite{he2026hgpo} and
GraphGPO~\cite{cheng2026graphgpo} add step-level terms from groups or graphs of recurring states; these
terms are silent where no state recurs and still come from the outcomes. Process reward models (PRMs)
score steps directly, but trained PRMs need step labels in the target
domain~\cite{lightman2024verify,wang2024mathshepherd} and transfer poorly beyond their training
distribution~\cite{zheng2024processbench,zhang2025prmlessons}; LLM judges are slow and costly and return text that has to be
parsed; and hand-written rules encode the form of a good step, not whether it serves the task.

\begin{center}
\emph{Every decision, in any task, gets its own credit:\\the last mile of agentic RL is now closed by decision models.}
\end{center}

We find that step credit need not be learned: it can be asked for. A \emph{general-purpose decision
model}~\cite{typesafe2026jev} answers a batch of yes/no questions about a state in one call and returns a probability
for each. Used zero-shot as a PRM, it needs one question text per environment and no step labels, no fine-tuning and
no critic. The same model yields step credit for fifty-step household episodes and for shopping episodes, with
three RL methods, at a small cost; we call this recipe \ours{} (decision-model PRM; \cref{fig:framework}). Agents trained with it
learn in far fewer iterations than those trained with GRPO and reach the goal in fewer steps. Our contributions:
\begin{itemize}
\item \textbf{A zero-shot decision model as a general-purpose PRM.} Nothing is trained besides the policy: no value
model, no PRM fitted to step labels. One decision model, with one question text per environment, scores
household and shopping episodes (\cref{sec:method}).
\item \textbf{\ours{}: plug-in step credit for any policy-gradient method.} The scores are standardised within the rollouts of a
task and added to the advantage after the method's own normalisation, so \ours{} plugs into any policy-gradient
method that computes an advantage; we add it to GRPO, GiGPO and GraphGPO. A check made before any update shows
what each step should be compared with (\cref{sec:method}).
\item \textbf{Large gains, beyond critic-based and group-based methods.}\ContribResults{} (\cref{sec:results}).
\item \textbf{Better agent behaviour.} Trained agents explore instead of circling and reach the goal in fewer steps
than GRPO's, whereas a rule in place of the decision model is gamed by actions that do nothing for the task
(\cref{sec:qual,sec:abl}).
\end{itemize}

\section{Method}
\label{sec:method}

\paragraph{Setting.} For a task $x$, the policy $\pi_\theta$ samples a group of $G$ rollouts. Rollout $i$
consists of steps $s_{i,1},\dots,s_{i,K_i}$; for an agent, a step is one turn (a model output and the observation
that follows it). A verifier returns an outcome $R_i$, and the RL method in use turns the outcomes of the group into
an advantage $A^{\text{RL}}_{i,k}$ for the tokens of each step. GRPO gives every step of rollout $i$ the same
value,
\begin{equation}
A^{\text{RL}}_{i,k} = \frac{R_i - \operatorname{mean}_j R_j}{\operatorname{std}_j R_j},
\end{equation}
which carries no signal about the task when all outcomes of the group are equal;\footnote{In the trainer we use, a 0.1 penalty for an output that is not a valid action enters the reward of
its step, so such a group still separates valid from invalid outputs.} GiGPO~\cite{feng2025gigpo} and GraphGPO~\cite{cheng2026graphgpo} add a step-level term computed
from the outcomes of rollouts that pass through the same state.

\paragraph{The decision model as a zero-shot PRM.} We use Jev~1.13~\cite{typesafe2026jev}, a general-purpose
decision model served through an API, as provided: we do not know its training data, do not tune it, and give it
no step labels. A request carries a state -- the task, the first observation, and the rollout's log as
``action $\rightarrow$ observation'' lines -- and one question per step: \emph{considering the task and everything
earlier in the log, did this step make progress toward completing the task?} The reply is one probability
$p_{i,k}\in[0,1]$ per step, with no free text to parse; scoring a fifty-step episode takes a few seconds. Each environment has one question text, fixed before the runs of \cref{tab:main}
(\cref{app:request}); the ALFWorld text states what counts (a useful new move) and what does not (no
effect, a repeat, a return to a searched place without a reason, an object the task does not need). The last line
of the log tells the decision model whether the episode succeeded, an outcome that is known during training anyway; the scores are
therefore not independent of the outcome, and what they add is its attribution to steps.

\paragraph{Step credit.} The advantage of the tokens of step $k$ becomes
$A_{i,k} = A^{\text{RL}}_{i,k} + w\, c_{i,k}$, with the step credit
\begin{equation}
c_{i,k} = \frac{p_{i,k} - \mu_x}{\max(\sigma_x, \sigma_{\min})},
\label{eq:group}
\end{equation}
where $\mu_x$ and $\sigma_x$ are the mean and standard deviation of the scores of \emph{all} steps of the $G$
rollouts of task $x$, and $\sigma_{\min}=0.1$ keeps the credits of a group with near-identical scores from being inflated.
The credit is added after the method's own normalisation, so both terms are of unit scale, and it is defined in
every group, including those in which all rollouts failed (scored episodes: \cref{app:cases}). A weight
$w=0$ recovers the underlying method. Nothing in \cref{eq:group} refers to how $A^{\text{RL}}$ was computed, so
the credit can be added to the advantage of any policy-gradient method; we add it to GRPO's, GiGPO's and
GraphGPO's. Like these methods, \ours{} is critic-free: where an actor--critic method fits a value
model to obtain a step-level signal, \ours{} queries a fixed model.

\paragraph{What a step is compared with: a check before any update.} \Cref{eq:group} compares a step with all
steps of the group, so a kind of action whose instances score above the group mean is pushed up as a whole, in
every state. In ALFWorld this push is desirable: going somewhere new scores above looking around. It is harmful
when the actions that complete a task are the ones an untrained agent performs badly. In WebShop the agent succeeds
only by opening a product, selecting its options and buying it; the untrained agent selects wrong options and buys
wrong products, so selecting options and buying score low while searching or going back scores far above the group mean, and the
credit of \cref{eq:group} teaches the agent to browse and to buy less. The two cases can be told apart on the
first iteration, before any update, from the mean score of each kind of action (\cref{tab:kinds} in the
appendix): in ALFWorld every kind that completes a task is above the mean of all steps; in WebShop selecting an
option and buying are below it, and searching and returning to the search page are the two highest. In the second
case we remove the push and compare like with like,
\begin{equation}
c_{i,k} = \frac{p_{i,k} - \mu_{x,\kappa(i,k)}}{\max(\sigma_x, \sigma_{\min})},
\label{eq:kind}
\end{equation}
where $\mu_{x,\kappa}$ is the mean score of the steps of kind $\kappa$ in the group. The kind of a step is the pair
(type of the page it is taken from, type of action): a search; on a results page, opening a product, turning the
page or going back; on a product page, selecting an option, opening a sub-page, buying, or leaving by either
button; and the corresponding kinds on a product's sub-pages. A step that had no effect -- a click on something
that is not there, or a repeated click -- counts as a step of the most frequent kind of its page (option clicks on a product
page), whatever its action, so that it is compared with the real steps of that kind. In every group the credits of
a kind sum to zero, so no kind of action is favoured as a whole: the credit only ranks the instances of a kind,
such as this product against the other products opened for the task, or a requested option against another value
or against a click that did nothing. The choice among kinds is left to the outcome term.

\paragraph{What the credit optimises.} With $\mu_x$ of \cref{eq:group} as the baseline, the added term is a
policy gradient with the step scores as rewards and a discount of zero, scaled by $w/\max(\sigma_x,\sigma_{\min})$: each action is
credited for its own step only. It is an auxiliary objective, not a potential-based shaping
term~\cite{ng1999shaping}, and can change what the policy converges to. In agent episodes it stays aligned with
the task: the question asks about progress, so a step that repeats an earlier one or does nothing scores low.
Under \cref{eq:kind} the baseline depends on the kind of action, and the term only ranks actions within a kind.
The credit is part of every update: a run that cannot reach the decision model waits and then stops
(\cref{app:impl}).

\begin{table}[t]
\centering\small
\setlength{\tabcolsep}{4.8pt}
\begin{tabular}{lcccccccccc}
\toprule
 & \multicolumn{8}{c}{ALFWorld success (\%)} & \multicolumn{2}{c}{WebShop (\%)} \\
\cmidrule(lr){2-9}\cmidrule(lr){10-11}
Method & Pick & Look & Clean & Heat & Cool & Pick2 & All & Unseen & Score & Succ. \\
\midrule
{\color{cited}GPT-4o} & {\color{cited}75.3} & {\color{cited}60.8} & {\color{cited}31.2} & {\color{cited}56.7} & {\color{cited}21.6} & {\color{cited}49.8} & {\color{cited}48.0} & {\color{cited}46.0$^{a}$} & {\color{cited}31.8} & {\color{cited}23.7} \\
{\color{cited}Gemini-2.5-Pro} & {\color{cited}92.8} & {\color{cited}63.3} & {\color{cited}62.1} & {\color{cited}69.0} & {\color{cited}26.6} & {\color{cited}58.7} & {\color{cited}60.3} & {\color{cited}50.5$^{a}$} & {\color{cited}42.5} & {\color{cited}35.9} \\
\midrule
Qwen2.5-1.5B-Instruct & 13.4 & 5.9 & 6.7 & 0.0 & 5.6 & 8.3 & 7.6 & 4.8 & 22.5 & 4.4 \\
{\color{cited}PPO} & {\color{cited}64.8} & {\color{cited}40.5} & {\color{cited}57.1} & {\color{cited}60.6} & {\color{cited}46.4} & {\color{cited}47.4} & {\color{cited}54.4} & {\color{cited}--} & {\color{cited}73.8} & {\color{cited}51.5} \\
{\color{cited}RLOO} & {\color{cited}88.3} & {\color{cited}52.8} & {\color{cited}71.0} & {\color{cited}62.8} & {\color{cited}66.4} & {\color{cited}56.9} & {\color{cited}69.7} & {\color{cited}68.7$^{a}$} & {\color{cited}73.9} & {\color{cited}52.1} \\
GRPO & 78.6 & 38.2 & 74.7 & 74.7 & 61.4 & 49.4 & 66.3 & 63.7 & 77.5 & 59.1 \\
GiGPO & 90.5 & 57.8 & 92.0 & 94.7 & 86.1 & 81.1 & 86.3 & 86.6 & 81.5 & 66.1 \\
{\color{cited}HGPO} & {\color{cited}--} & {\color{cited}--} & {\color{cited}--} & {\color{cited}--} & {\color{cited}--} & {\color{cited}--} & {\color{cited}\underline{92.8}} & {\color{cited}90.2} & {\color{cited}85.6} & {\color{cited}71.5} \\
GraphGPO & 83.0 & 70.6 & \textbf{98.7} & \textbf{100.0} & 88.8 & \textbf{86.7} & 88.6 & 88.3 & 84.4 & \underline{73.8} \\
\midrule
\rowcolor{oursbg}\ours{}$_{\text{GRPO}}$ (ours) & 97.6 & \underline{91.2} & 95.1 & 94.0 & 89.1 & 66.7 & 90.0 & 92.9 & 84.6 & 72.2 \\
\rowcolor{oursbg}\quad{\scriptsize\emph{gain over GRPO}} & {\scriptsize {\color{gain}$+$19.0}} & {\scriptsize {\color{gain}$+$52.9}} & {\scriptsize {\color{gain}$+$20.4}} & {\scriptsize {\color{gain}$+$19.3}} & {\scriptsize {\color{gain}$+$27.7}} & {\scriptsize {\color{gain}$+$17.2}} & {\scriptsize {\color{gain}$+$23.7}} & {\scriptsize {\color{gain}$+$29.2}} & {\scriptsize {\color{gain}$+$7.1}} & {\scriptsize {\color{gain}$+$13.1}} \\
\rowcolor{oursbg}\ours{}$_{\text{GiGPO}}$ (ours) & \underline{99.4} & \textbf{92.2} & 92.0 & 92.7 & \textbf{95.1} & 67.8 & 91.3 & \underline{93.5} & \underline{85.7} & 69.6 \\
\rowcolor{oursbg}\quad{\scriptsize\emph{gain over GiGPO}} & {\scriptsize {\color{gain}$+$8.9}} & {\scriptsize {\color{gain}$+$34.3}} & {\scriptsize 0.0} & {\scriptsize {\color{loss}$-$2.0}} & {\scriptsize {\color{gain}$+$9.0}} & {\scriptsize {\color{loss}$-$13.3}} & {\scriptsize {\color{gain}$+$4.9}} & {\scriptsize {\color{gain}$+$6.9}} & {\scriptsize {\color{gain}$+$4.2}} & {\scriptsize {\color{gain}$+$3.5}} \\
\rowcolor{oursbg}\ours{}$_{\text{GraphGPO}}$ (ours) & \textbf{100.0} & 85.3 & \underline{96.0} & \underline{97.3} & \underline{94.0} & \underline{82.8} & \textbf{94.0} & \textbf{93.6} & \textbf{86.3} & \textbf{75.9} \\
\rowcolor{oursbg}\quad{\scriptsize\emph{gain over GraphGPO}} & {\scriptsize {\color{gain}$+$17.0}} & {\scriptsize {\color{gain}$+$14.7}} & {\scriptsize {\color{loss}$-$2.7}} & {\scriptsize {\color{loss}$-$2.7}} & {\scriptsize {\color{gain}$+$5.2}} & {\scriptsize {\color{loss}$-$3.9}} & {\scriptsize {\color{gain}$+$5.5}} & {\scriptsize {\color{gain}$+$5.2}} & {\scriptsize {\color{gain}$+$1.9}} & {\scriptsize {\color{gain}$+$2.1}} \\
\bottomrule
\end{tabular}

\caption{ALFWorld (Pick--All: seen rooms) and WebShop: prompted closed-source models (top) and Qwen2.5-1.5B-Instruct, untrained (third row) and after RL post-training (final checkpoints, pass@1, means over runs).
\emph{Black}: our runs, on the same evaluation episodes (standard deviations: \cref{tab:runs}); shaded: \ours{}. {\color{cited}\emph{Grey}: from the cited papers}
(\TabGreySources{}; --: not reported there). \emph{Gain}: over the method it is added to, before rounding. Bold: best; underlined:
second.}
\label{tab:main}
\end{table}

\section{Experiments}
\label{sec:exp}

\subsection{Experimental Setup}
\label{sec:setup}

\paragraph{Benchmarks.} ALFWorld~\cite{shridhar2021alfworld}, built on TextWorld~\cite{cote2018textworld}, is a text version of ALFRED's household
tasks~\cite{shridhar2020alfred}: six task types in rooms whose
contents the agent must discover by acting, tested in rooms seen and unseen in training. WebShop~\cite{yao2022webshop}
simulates a shopping site: an instruction names a product, attributes, option values and a price limit, and the
agent searches, opens products, selects options and buys. The reward is the binary outcome of the episode (WebShop also reports a task score).

\paragraph{Training.} We follow the verl-agent setup of Feng et al.~\cite{feng2025gigpo}, built on
verl~\cite{sheng2025hybridflow} and vLLM~\cite{kwon2023vllm}, and keep its prompts, parser and settings (\cref{app:setup}). The policy, Qwen2.5-1.5B-Instruct~\cite{qwen2024qwen25}, is trained for 150 iterations of 16 tasks $\times$ 8
rollouts.

\paragraph{Methods.} We add \ours{} to GRPO, GiGPO and GraphGPO (\ours{}$_{\text{GRPO}}$,
\ours{}$_{\text{GiGPO}}$ and \ours{}$_{\text{GraphGPO}}$). All three use the same decision model and question text per environment,
with the pooled baseline of \cref{eq:group} and $w=1$ on ALFWorld, and the per-kind baseline of
\cref{eq:kind} and $w=0.3$ on WebShop. For fairness we reproduce each base method with its released code and add
\ours{} inside it: the verl-agent codebase of
Feng et al.~\cite{feng2025gigpo} for GRPO and GiGPO, the release of Cheng et al.~\cite{cheng2026graphgpo} for GraphGPO
(\cref{app:setup}). GiGPO denotes its variant that divides the step-level term by its standard deviation (w/ std; on WebShop the release default is w/o std: \cref{app:setup}).

\paragraph{Evaluation.} All final checkpoints are evaluated on the same episodes, one attempt each (pass@1): 420 ALFWorld episodes in seen
rooms, 420 in unseen rooms, and 500 WebShop episodes. On ALFWorld, each of GRPO, GiGPO and \ours{}$_{\text{GRPO}}$ is trained three
times, on three sequences of training games shared by the methods, and we report means. On WebShop
we use the standard split and add a split whose test goals are never used in training. Further discussion, runs per method,
standard deviations and paired comparisons are in \cref{app:discussion,app:details,app:additional}.

\subsection{Main Results}
\label{sec:results}

\begin{figure}[t]
\centering
\includegraphics[width=\textwidth]{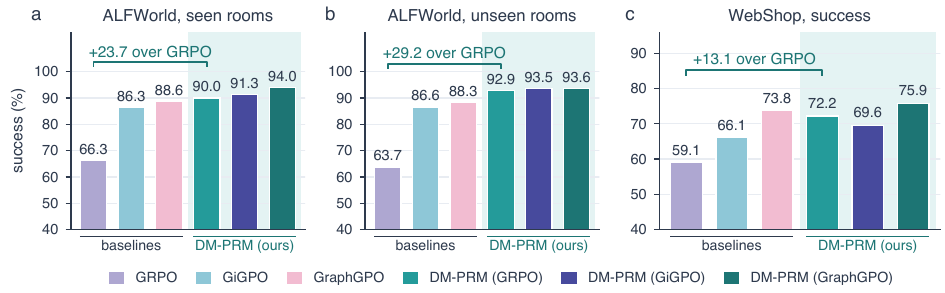}
\caption{Added to GRPO, \ours{} lifts success by \AlfJevOverGrpoId{}, \AlfJevOverGrpoOod{} and \WsJevOverGrpo{} points (brackets). Shaded: \ours{} on the named method. Our own runs on the same evaluation
episodes; means over runs (the rows of \cref{tab:main}). (a,~b) ALFWorld. (c) WebShop.}
\label{fig:summary}
\end{figure}

\begin{figure}[t]
\centering
\includegraphics[width=\textwidth]{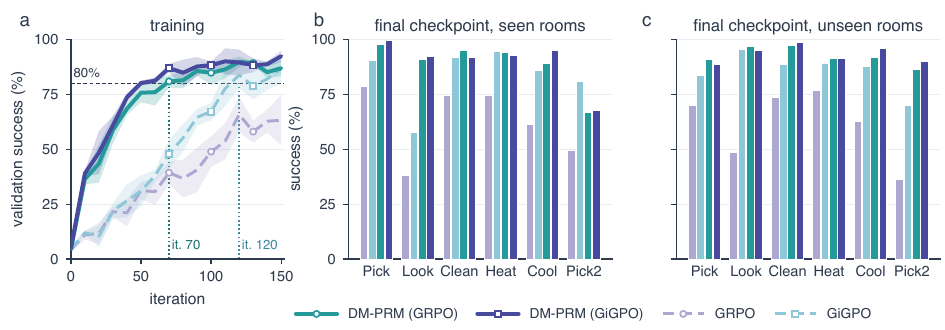}
\caption{ALFWorld. (a) \ours{}$_{\text{GRPO}}$ reaches 80\% validation success at iteration \AlfValJevIterEighty{}, GiGPO at
\AlfValGigpoIterEighty{}, GRPO not within 150 (the trainer's validation, 128 episodes every 10 iterations; means over runs, band: one standard deviation). (b, c) Final
checkpoints by task type on the 420 + 420 evaluation episodes: \ours{}$_{\text{GRPO}}$ is above GRPO on every type (colours as in a; standard
deviations for b: \cref{tab:runs}).}
\label{fig:alf}
\end{figure}

\begin{figure}[t]
\centering
\includegraphics[width=\textwidth]{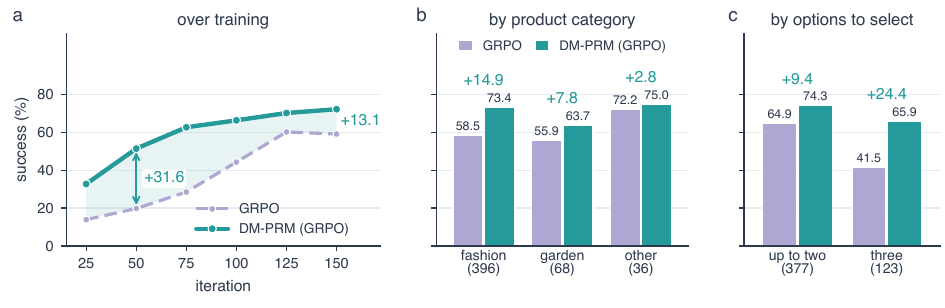}
\caption{WebShop, \ours{}$_{\text{GRPO}}$ against GRPO on the 500 evaluation episodes (means over runs). (a) Checkpoints at
iterations 25 to 150: \ours{} is ahead at every one. (b,~c) Final checkpoints by product category (other: beauty, electronics,
grocery) and by the number of options a goal asks for (episodes in brackets).}
\label{fig:ws}
\end{figure}

\paragraph{ALFWorld: large and consistent gains over GRPO.} After 150 iterations, the credit lifts the success of GRPO by \AlfLiftSeenWords{} points on seen rooms and by
\AlfLiftUnseenWords{} on unseen rooms (\cref{fig:summary}, \cref{tab:main}). The gain is positive on \AlfJevGrpoTypesAhead{} one of the six task
types, on seen and on unseen rooms (\cref{fig:alf}b, c), and in each of the three runs. \ours{}$_{\text{GRPO}}$ also overtakes GiGPO, whose step-level term needs rollouts that pass
through identical states: paired on the same episodes, it is ahead on seen rooms and clearly ahead on unseen
rooms, where it leads on \AlfJevGigpoTypesAheadOod{} task types (\cref{fig:alf}c, \cref{app:runs}).
On seen rooms, the gain over both GRPO and GiGPO is largest in examining an object under a lamp (Look).

\paragraph{ALFWorld: faster learning from the start.} In each of the three runs, the validation curve of \ours{}$_{\text{GRPO}}$ is
above those of GRPO and GiGPO from iteration 10 through 110 (\cref{fig:alf}a and \cref{fig:app_curves_sets}). Its mean reaches 80\% \AlfIterWords{} and stays there, whereas GRPO does not
reach it within 150 iterations. The reason is visible in the first iteration: in most task groups no rollout succeeds, so the
outcome tells GRPO nothing about the task, while the credit still separates the useful steps of these rollouts from the
rest.

\paragraph{A plug-in for any policy-gradient method.} \ours{} is not tied to GRPO: it runs unchanged on methods that already
have a step-level term (gain rows of \cref{tab:main}).\PlugSentences{}
The gain is largest on GRPO, which has no step-level signal of its own. The terms of GiGPO and GraphGPO come
from rollouts that share a state, whereas ours comes from a judgement of the step itself and exists even when no rollout succeeds.

\begin{figure}[t]
\centering
\includegraphics[width=\textwidth]{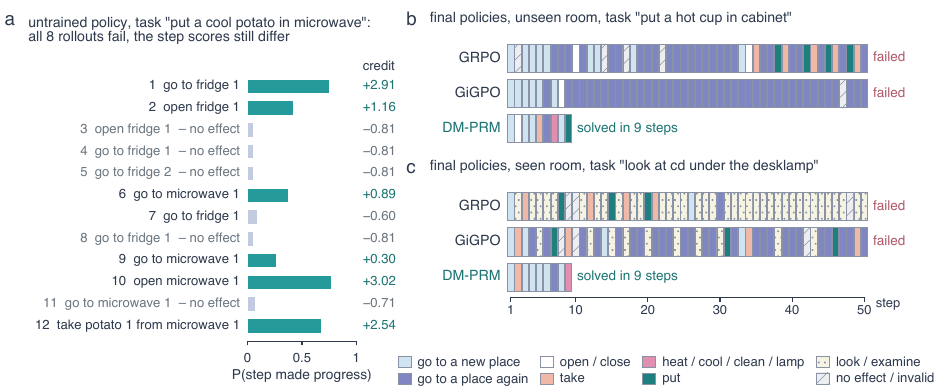}
\caption{Case study. (a) All eight rollouts of this task fail, yet the decision model tells their steps apart: its probability that a
step made progress (bars; teal: positive credit) and the resulting credit (numbers) on the first 12 of 50 steps of one rollout of the untrained policy. (b, c) Final policies of one run per method on the same evaluation
episodes, one tile per step: \ours{}$_{\text{GRPO}}$ solves both tasks in nine steps; GRPO and GiGPO use up the
50 steps.}
\label{fig:case}
\end{figure}

\paragraph{One decision model for both benchmarks.} Used zero-shot, without labels or training, the same decision
model scores household and shopping episodes; between the two only the
question text, the log format, the weight and the baseline of the credit differ (\cref{sec:method}). Together with the plug-in
results, this makes our process rewards general-purpose: one decision model serves both benchmarks and three RL
methods unchanged, and only the per-environment settings above change.

\paragraph{WebShop: a large gain over GRPO and faster learning.}\WsEndSentence{} It learns faster and stays ahead of
GRPO at every checkpoint, on held-out test goals as well (\cref{app:ws}); in \cref{fig:ws}a,\WsHalfWayClause{}\WsSplitSentence{}\WsOthersSentence{}

\paragraph{Beyond critic-based and group-based methods.} In \cref{tab:main} (grey cells: the papers' numbers, \cref{app:runs}),\PpoSentenceB{} Added to GRPO, which has no step-level term, \ours{} is above GiGPO on
ALFWorld\AheadWsClause{}.\BestRowsClause{} GiGPO, HGPO~\cite{he2026hgpo} and GraphGPO build their step-level terms from states that recur across rollouts;
\ours{} needs no such structure, only a question.

\subsection{Case Study}
\label{sec:case}

\paragraph{Credit when every rollout fails.} Why does the credit help from the first iterations? \Cref{fig:case}a shows a rollout from a group in which all
eight rollouts fail, so the outcome tells their steps apart only by whether they are valid actions. The decision
model separates them: reaching and opening the fridge and the microwave and taking the potato score high; the attempts that
did nothing score near zero. This holds over all such groups of iteration 1: steps that go somewhere new, open or
handle an object are a fifth of the steps but nearly all of the top tenth ranked by credit
(\cref{app:cases}). The credit thus gives a signal about the task exactly where the outcome gives none.

\paragraph{The trained agent searches; the baselines circle.} In \cref{fig:case}b the task is to put a hot
cup in a cabinet, in a room never seen in training. The policy trained with the credit opens the microwave, looks
into places it has not yet searched until it finds a cup, heats it and puts it away: nine steps. GiGPO's policy visits
six places and then moves among the same six for the rest of the fifty steps; GRPO's finds a cup late and carries
it around without heating it. In \cref{fig:case}c all three policies pick up the CD at once. Ours then walks
on through the room until it finds the lamp and turns it on, GRPO's examines the CD where it stands, and GiGPO's
keeps walking between the same four places until the episode ends.

\subsection{Qualitative Analysis}
\label{sec:qual}

\begin{figure}[t]
\centering
\includegraphics[width=0.5\textwidth]{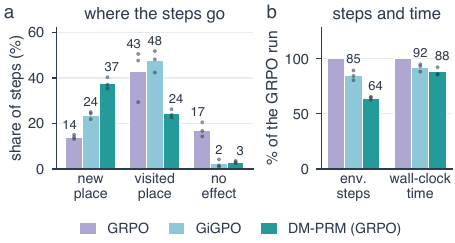}
\caption{(a) With \ours{} the final ALFWorld policies go to new places while those of GRPO and GiGPO return to
visited ones (share of steps on the evaluation episodes). (b) A training run then takes fewer steps and less time than the GRPO run on
the same games, scoring included. Bars: means over three runs; dots: single runs.}
\label{fig:behcost}
\end{figure}

\paragraph{The agents explore instead of circling.} The pattern of the case study holds across the evaluation episodes. About three times as many
evaluation episodes are solved only by our policy as only by GiGPO's (\cref{app:qualalf}). The final
policies of GRPO and GiGPO return to places they have already visited far more often than they go to new ones;
with the credit this is reversed (\cref{fig:behcost}a), and against GRPO the goal is reached in fewer steps,
not only more often.

\paragraph{A large gain at a small cost.} Because episodes end sooner, an ALFWorld run with the credit uses
over a third fewer environment steps than the GRPO run on the same games and finishes in less wall-clock time
than GRPO and GiGPO, although it waits for the decision model at every iteration (\cref{fig:behcost}b). No value
model is trained and no additional GPU is used; scoring takes under 2\% of the training time, and its API
calls cost under \$20 per run (\cref{tab:cost} in the appendix), an amount that scales with the size of the run, not with the size of the policy.

\paragraph{On WebShop the agents buy the right product.} Sorting the evaluation episodes of the final
checkpoints by how they end (\cref{tab:wsloss} in the appendix) shows where the gain over GRPO comes from:
with the credit, about half as many episodes end with a product other than the goal's, and fewer with the goal's product but wrong options. GiGPO buys another product far more often than our policy. Added to GraphGPO, the credit lowers this share
as well. On GRPO and GraphGPO alike, the credit corrects the choice of product: the step that decides a WebShop episode and that the outcome alone cannot single out among its many steps.

\paragraph{The scores judge the step, not the episode.} The decision model is told how the episode ended, so its scores
could merely echo the outcome. On ALFWorld they do not: the kind of step explains several times more of the variance of the
scores than the outcome does, almost all of the variance of the credit lies within episodes, and
an action without effect scores the same in won and lost episodes (\cref{app:val}).

\paragraph{Far cheaper than an LLM judge.}\label{sec:scope} A large language model prompted as a judge could also score every step.
On samples of four error-location benchmarks (ProcessBench~\cite{zheng2024processbench}, MR-Ben~\cite{zeng2024mrben},
BIG-Bench Mistake~\cite{tyen2024bbmistake}, Who\&When~\cite{zhang2025whowhen}), a frontier LLM
judge costs tens to hundreds of times more than the decision model (\cref{tab:bench} in the appendix), is far slower
and returns text that has to be parsed, a heavy cost for a loop that scores every step of every rollout.

\subsection{Ablation Study}
\label{sec:abl}

\begin{table}[t]
\centering\small
\setlength{\tabcolsep}{3.2pt}
\begin{tabular}{lcccccc}
\toprule
Configuration & \shortstack{step\vphantom{dp}\\credit\vphantom{dp}} & \shortstack{decision\vphantom{dp}\\model\vphantom{dp}} & \shortstack{state\vphantom{dp}\\groups\vphantom{dp}} & Val. & Seen & Unseen \\
\midrule
GRPO & $\times$ & $\times$ & $\times$ & 32.0 & 37.1 & 26.4 \\
GiGPO & $\times$ & $\times$ & \checkmark & 40.6 & 44.3 & 36.7 \\
GRPO + rule credit & \checkmark & $\times$ & $\times$ & 11.7 & 11.4 & 5.2 \\
\midrule
\rowcolor{oursbg}\ours{}$_{\text{GRPO}}$ & \checkmark & \checkmark & $\times$ & 71.1 & 72.1 & 82.9 \\
\rowcolor{oursbg}\quad weight $w{=}0.3$ & \checkmark & \checkmark & $\times$ & \underline{78.1} & \underline{77.1} & \underline{83.6} \\
\rowcolor{oursbg}\quad reworded question & \checkmark & \checkmark & $\times$ & 71.9 & 72.9 & 77.1 \\
\rowcolor{oursbg}\ours{}$_{\text{GiGPO}}$ & \checkmark & \checkmark & \checkmark & \textbf{82.8} & \textbf{80.7} & \textbf{87.1} \\
\bottomrule
\end{tabular}

\caption{Ablation on ALFWorld (iteration 60, same training games; success, \%; one run per row).
Components: a step credit, the decision model as its source (third row: a rule), the state groups of GiGPO.
\emph{Val.}: validation during training; \emph{Seen}, \emph{Unseen}: the 420 + 420 evaluation episodes. Bold: best; underlined: second.}
\label{tab:abl}
\end{table}

\begin{figure}[t]
\centering
\includegraphics[width=0.5\textwidth]{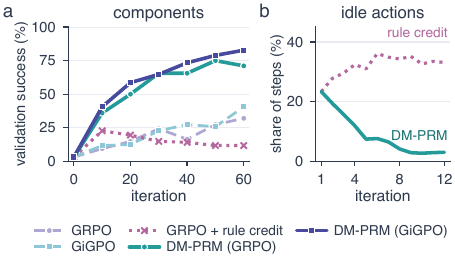}
\caption{Ablation over training (runs of \cref{tab:abl}). (a) The decision model's credit lifts validation
success from the first iterations, on GRPO and on GiGPO; a rule credit helps at iteration 10 and then falls below
GRPO. (b) Under the rule, idle actions (look, inventory, examine) become more frequent; with \ours{}$_{\text{GRPO}}$ they almost vanish (23\% of steps to 3\%).}
\label{fig:abl}
\end{figure}

\paragraph{Components.} Would any step credit do? It works only if it comes from the decision model (\cref{tab:abl}, \cref{fig:abl}a):
with a rule as its source, the run ends near the untrained policy. The decision
model's credit alone lifts GRPO far above GiGPO, and the two signals combine: \ours{}$_{\text{GiGPO}}$ is the best row on
seen and unseen rooms.

\paragraph{A rule in place of the decision model.} As a control, we replace the decision model's score by
a rule: 0 if the step had no effect, repeated an earlier step with the same result or was not a valid action, and
1 otherwise. Within eight updates, the rule credit removes actions without
effect faster than the decision model's does, but the share of \texttt{look} / \texttt{inventory} /
\texttt{examine} \emph{rises}: these actions have an effect and so satisfy the rule, yet do nothing for the task
(\cref{fig:abl}b). The decision model scores such steps near zero, and they almost disappear. Under the rule,
validation success first rises and then falls: the rule is gamed, the decision model is not.

\paragraph{Weight and question.} On ALFWorld, a weight of 0.3 learns more slowly in the first 20 iterations and is ahead at iteration 60; on
WebShop the weight matters more (\cref{app:ws}). A reworded question that also credits returning to a known
place changes little on seen rooms and is lower on unseen ones; we keep the first wording.

\paragraph{The baseline of the credit.} The scores alone are not a credit: the baseline they are compared with
decides what is reinforced. On ALFWorld the pooled baseline of \cref{eq:group} works because the actions that
complete a task score above the average step. On WebShop it misleads: an untrained agent buys wrong products, so
purchases score low and searching scores high, and with one baseline for all steps the agent learns to browse and to
buy less (\cref{app:ws}). Comparing each step only with the steps of its own kind
(\cref{eq:kind}) removes this push and gives the WebShop results above.

\section{Related Work}
\label{sec:related}

\paragraph{Process reward models.} PRMs, introduced to verify or rerank reasoning chains, were trained on human
step labels~\cite{lightman2024verify} or on labels from rollouts~\cite{wang2024mathshepherd}.
ProcessBench~\cite{zheng2024processbench} finds trained PRMs below prompted LLMs outside their training
distribution~\cite[see also][]{zhang2025prmlessons}. PRMs have since been trained for agents, on annotated web
trajectories (Web-Shepherd~\cite{web_shepherd}) or from the promise and progress of each
step (AgentPRM~\cite{AgentPRM}), to verify or guide an agent's actions. We use step scores during RL
training of agents, as a step credit added to the advantage, and obtain them zero-shot from a general-purpose decision
model that we do not train and that returns probabilities, not text.

\paragraph{Credit assignment in RL post-training of agents.} GRPO~\cite{shao2024deepseekmath} and its
successors~\cite{yu2025dapo} credit a response by its outcome relative to its group;
PPO~\cite{schulman2017ppo} with GAE~\cite{schulman2016gae} needs a learned critic. For multi-turn agents,
GiGPO~\cite{feng2025gigpo} adds a step-level advantage from groups of identical states, and a growing family
refines the grouping or replaces it with a graph over the visited
states~\cite{he2026hgpo,zhu2026gagpo,cheng2026graphgpo,yao2026graft,zhu2026graphhca}. All of these derive the step
signal from the outcomes of the group; our step credit comes from a judgement of the step itself and can be added to
any of them. Several of these, and further concurrent
work~\cite{fan2026progpo,jin2026himac,zheng2026pgpo,yan2026gatedbepo}, are preprints; we compare with the
peer-reviewed methods.
Other work adds signals that are not outcomes, such as predicting the next
observation~\cite{li2026tapo,wang2026roleagent}. T$^3$~\cite{zou2026t3} and AREW~\cite{zou2026arew} truncate
rollouts or reweight advantages with task-specific progress signals in active-reasoning tasks; a general-purpose
decision model is a candidate source of such signals.

\section{Conclusion}

We introduce \ours{}, which uses a decision model, zero-shot, as a general-purpose process reward model: it asks one
yes/no question per step and adds the standardised answers to the advantage of any policy-gradient method, with no
step labels, no value model and no text to parse. \ConclResults{} One decision model serves two environments and
three RL methods unchanged: no credit model has to be learned for a task. The step credit is defined even when every
rollout of a task fails and the outcome is silent, and learning is faster: on WebShop, \ours{}$_{\text{GRPO}}$
reaches the final success of GRPO in half the iterations. On ALFWorld, trained agents explore instead of circling and
reach the goal in fewer steps than GRPO's, which shortens training; on WebShop, they buy the right product more often.
A rule in place of the decision model is gamed, and the baseline of the credit has to fit the environment. The recipe
is general-purpose in this sense: a new environment needs only a question text with a log format, a weight and, if
needed, a baseline per kind of action, and a new RL method needs only its own advantage. For language agents, step
credit need not be learned: it can be asked for.

\ours{} calls an external decision model during training, which makes a run depend on access to that service and
adds an API cost and a delay to every iteration. Both are small in our setting: scoring the 128 rollouts of an
iteration takes about ten seconds and costs about \$0.12 (\cref{app:cost}), though they grow with the number
and length of the rollouts. The decision model is closed and used as provided, and our experiments cover two agent
benchmarks with one policy model of 1.5B parameters; a new environment needs its own question text and may need a
baseline per kind (\cref{eq:kind}; \cref{app:request,app:ws}).

\clearpage
\pdfbookmark[1]{References}{references}
\bibliographystyle{unsrtcase}
\bibliography{main}

\clearpage
\appendix
\crefalias{section}{appendix}\crefalias{subsection}{subappendix}\crefalias{subsubsection}{subsubappendix}
\section{Extended Discussion}
\label{app:discussion}

\paragraph{A general-purpose decision model is a general-purpose process reward model.} The main finding of this
paper is that step credit for language agents does not have to be learned. Learned sources of step credit are tied
to a task: a value model is fitted to its outcomes, a PRM to its step labels. A general-purpose decision model, used
as provided with one question text per environment, serves instead as a general-purpose process reward model across
tasks and RL methods, and it is cheap enough to score every step of every rollout inside the training loop. It lifts GRPO by \AlfLiftSeenWords{} points on ALFWorld and by \WsLiftWords{} on WebShop
(\cref{tab:main}), and on both benchmarks the best overall results of our comparison are obtained with \ours{},
ahead of the critic-based PPO and of the group-based GiGPO, HGPO and GraphGPO.

\paragraph{Credit where the outcome is silent.} Outcome-based methods learn from differences between the
rollouts of a task. When every rollout fails, as in most task groups at the start of training, the outcome carries no
information about the task, yet the decision model still tells the steps of
these rollouts apart (\cref{fig:case}a and \cref{fig:qual_scorer}): it rates a step that reaches a new
place or picks up a required object far above one that has no effect. This is where the advantage of \ours{}
begins: it learns faster early in training on both benchmarks (\cref{fig:alf}a and \cref{fig:ws}a) and
keeps its lead over GRPO until the end.

\paragraph{Why not a learned critic, hand-written rules or an LLM judge?} A value model is a second network that must be held in memory and fitted alongside the
policy from the same sparse outcomes: in the setting of \cref{tab:main}, PPO with its critic ends far below
the group-based methods and below every \ours{} row on ALFWorld. Hand-written rules need no training, but they
describe the form of a good step, not its use: in our ablation, a rule that credits every action with an effect
is satisfied by looking around, and the policy learns exactly that (\cref{sec:abl}). A large language model
prompted as a judge is far more expensive and far slower
(\cref{app:judge}), a heavy cost for a loop that scores every step of
every rollout. The decision model combines the strengths of rules and judges: like a rule, it needs neither training
nor labels; like a judge, it reads the task and the history of the episode before scoring a step; and its cost is
small next to that of training (\cref{app:cost}).

\paragraph{Plug-and-play for any policy-gradient method.} \ours{} is general-purpose and plug-and-play: it applies to any
policy-gradient method that computes an advantage, without changing that method. The credit is added to the
advantage after the method's own normalisation; the loss and the optimiser stay as they are
(\cref{sec:method}). Added in this way to GRPO, GiGPO and GraphGPO, it improves all three on both benchmarks
(\cref{tab:main}): most on GRPO, which has no step-level signal of its own, but also on GiGPO and GraphGPO,
whose step-level terms compare rollouts that pass through the same state. The two signals are complementary: one
compares rollouts; the other judges the step itself and exists even when no rollout succeeds.

\paragraph{The baseline decides what the credit teaches.} A second finding is that the scores of a decision model
are not yet a credit. When each step is compared with all steps of its task, whole kinds of action are favoured over
others. In ALFWorld this helps: the kinds of action that complete a task, such as going somewhere new, score above
the average step. In WebShop it misleads: an untrained agent buys wrong products, so browsing scores higher than
buying, and the agent learns to buy less. A check on the first iteration, before any update, tells the two cases
apart, and comparing each step only with steps of its own kind removes the unwanted preference
(\cref{sec:method}, \cref{tab:kinds}, \cref{app:ws}).

\paragraph{Better behaviour, not only higher success.} The credit also changes how the trained agents act: they go to
places they have not yet searched, whereas the baselines return to visited ones; they drop idle actions within a dozen iterations (\cref{fig:abl}b), reach the goal in fewer steps
than GRPO's agents and, on WebShop, buy the right product more often
(\cref{sec:qual}, \cref{app:qual}). Because episodes end sooner, an ALFWorld run with the
credit also takes less time than the GRPO run on the same games (\cref{app:cost}).

\paragraph{What a new setting needs.} A new environment needs a question text with a format for the episode log,
a weight and, if the first-iteration check calls for it, a baseline per kind of action; a new RL method needs only
its own advantage (\cref{app:algo}). Natural next steps are larger policies, longer episodes, other
questions about a step and other decision models.

\clearpage
\section{Algorithm Pseudocode}
\label{app:algo}

\Cref{alg:dmprm} gives one training iteration of \ours{}. The only additions to the trainer are the
scoring requests and the lines that turn the scores into a credit; the advantage, the loss and the optimiser
are those of the RL method in use. The credit enters every update: if the decision model cannot be reached, the run waits
and then stops (\cref{app:impl}).

\makeatletter
\let\dmprmState\State
\renewcommand{\State}{\dmprmState\parshape 2 \@totalleftmargin \linewidth \dimexpr\@totalleftmargin+\ALG@tlm\relax \dimexpr\linewidth-\ALG@tlm\relax}
\makeatother
\algrenewcommand\algorithmicindent{1em}
\begin{algorithm}[H]
\caption{\ours{}: one training iteration}
\label{alg:dmprm}
\begin{algorithmic}[1]
\Require policy $\pi_\theta$; RL method $\mathcal{E}$ (GRPO, GiGPO or GraphGPO); decision model $D$ with question text $q$; weight $w$; floor $\sigma_{\min}$; baseline: pooled (\cref{eq:group}) or per kind (\cref{eq:kind})
\For{each task $x$ of the batch}
  \State sample $G$ rollouts $\tau_1,\dots,\tau_G\sim\pi_\theta(\cdot\mid x)$; the verifier returns $R_1,\dots,R_G$
  \State $A^{\text{RL}}_{i,k}\gets\mathcal{E}\big(\{(\tau_i,R_i)\}_{i=1}^{G}\big)$ \Comment{the method's advantage and normalisation, unchanged}
  \For{$i=1,\dots,G$}
    \State $(p_{i,1},\dots,p_{i,K_i})\gets D\big(\text{log of }\tau_i,\;q\big)$ \Comment{one yes/no question per step}
  \EndFor
  \State $\sigma_x\gets$ standard deviation of $p_{i,k}$ over all steps of the $G$ rollouts
  \State $\mu_{i,k}\gets$ mean of $p$ over all steps of the group (pooled), or over its steps of the same kind as step $(i,k)$ (per kind)
  \State $c_{i,k}\gets(p_{i,k}-\mu_{i,k})\,/\max(\sigma_x,\sigma_{\min})$ \Comment{step credit}
  \State $A_{i,k}\gets A^{\text{RL}}_{i,k}+w\,c_{i,k}$ for the tokens of step $k$ of rollout $i$
\EndFor
\State update $\theta$ with the trainer's objective on the advantages $A$ \Comment{loss, KL term and optimiser unchanged}
\end{algorithmic}
\end{algorithm}

\begin{figure}[!ht]
\centering
\includegraphics[width=0.55\textwidth]{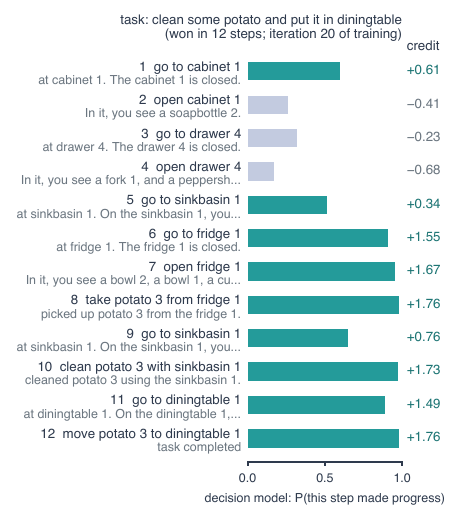}
\caption{One won ALFWorld episode from training with \ours{}$_{\text{GRPO}}$ (iteration 20). Bars: the decision model's probability that the
step made progress; right: the credit (teal bar: positive, grey: negative). The outcome gives every step the same advantage; the credit separates the
fruitless searches (steps 2--4) from the steps that found, cleaned and delivered the potato.}
\label{fig:example}
\end{figure}

\paragraph{Applying \ours{} to a new environment.} Since nothing is trained besides the policy, a new environment
needs only three things. (i) One question text and a format for the episode log
(\cref{app:request}). (ii) A check on the first iteration, before any update, that compares the mean score
of each kind of action with the mean of all steps (\cref{tab:kinds}). If the kinds of action that complete a task score
above the mean, as in ALFWorld, the pooled baseline of \cref{eq:group} pushes the agent towards them and is the
right choice. If they score below it, as selecting an option and buying do in WebShop, the baseline per kind of
\cref{eq:kind} removes the push and lets the credit rank only the instances of a kind. (iii) A weight: we use
$w=1$ with the pooled baseline and $w=0.3$ with the baseline per kind (\cref{app:ws}).
\Cref{fig:example} shows the scores and the resulting credit on one episode.

\begin{table}[!ht]
\centering\footnotesize
\setlength{\tabcolsep}{3pt}
\begin{tabular}{lr@{\hskip 8pt}lr}
\toprule
\multicolumn{2}{l}{ALFWorld (mean 0.15)} & \multicolumn{2}{@{}l}{WebShop (mean 0.20)} \\
\midrule
use a lamp & 0.45 & back to search & 0.47 \\
open & 0.40 & search & 0.46 \\
take & 0.37 & open a product & 0.25 \\
move / put & 0.37 & turn a page & 0.24 \\
heat / cool / clean & 0.35 & option click with an effect & 0.18 \\
go to & 0.26 & open a sub-page & 0.13 \\
close & 0.12 & buy & 0.08 \\
look / examine & 0.09 & no effect & 0.06 \\
no effect & 0.06 & & \\
\bottomrule
\end{tabular}
\caption{Mean score by kind of action in the first training iteration, before any update (ALFWorld: 5,819 steps
of one run; WebShop: 1,643 steps; ``mean'' is over all steps of the iteration; look /
examine includes inventory). In ALFWorld the kinds that complete a task are above the mean of all steps, so
comparing every step with all steps pulls the agent towards them. In WebShop selecting an option and buying are below the
mean of all steps, opening a product is slightly above it, and searching and going back are far above it.}
\label{tab:kinds}
\end{table}

\clearpage
\section{Experimental Details}
\label{app:details}

\subsection{Training and Evaluation Setup}
\label{app:setup}

\paragraph{Training.} All runs use the verl-agent trainer with the prompts (\cref{app:prompts}), parser and settings of
its examples, except where noted in \cref{tab:hyper} and below; environments, prompts and model outputs are in English.
\ours{} adds
the weight $w$, the floor $\sigma_{\min}$, the baseline of the credit and the question text of \cref{app:request}.
Neither $\sigma_{\min}$ nor the ALFWorld weight was tuned; the WebShop baseline and weight were chosen on short pilot runs
(\cref{app:ws}).

\begin{table}[!ht]
\centering\small
\setlength{\tabcolsep}{4pt}
\begin{tabular}{lcc}
\toprule
 & ALFWorld & WebShop \\
\midrule
policy model & \multicolumn{2}{c}{Qwen2.5-1.5B-Instruct} \\
iterations & \multicolumn{2}{c}{150} \\
tasks $\times$ rollouts per iteration & \multicolumn{2}{c}{16 $\times$ 8} \\
steps per episode (at most) & 50 & 15 \\
steps of history in the prompt & \multicolumn{2}{c}{2} \\
prompt limit (tokens) & 2,048 / 3,072$^{*}$ & 4,096 / 5,120$^{*}$ \\
response limit (tokens) & \multicolumn{2}{c}{512} \\
learning rate & \multicolumn{2}{c}{$10^{-6}$} \\
weight of the KL term & \multicolumn{2}{c}{0.01} \\
penalty for an invalid action & \multicolumn{2}{c}{0.1} \\
validation: episodes, every & 128, 10 it. & 128, 5 it. \\
\midrule
GiGPO: $\gamma$, step weight & \multicolumn{2}{c}{0.95, 1} \\
GiGPO: step-level term & w/ std & w/ std \\
GraphGPO: $\gamma$, step / episode weight & 0.10, 1 / 1 & 0.20, 1 / 1 \\
\midrule
\ours{}: weight $w$ & 1 & 0.3 \\
\ours{}: baseline of the credit & \cref{eq:group} & \cref{eq:kind} \\
\ours{}: floor $\sigma_{\min}$ & \multicolumn{2}{c}{0.1} \\
\bottomrule
\end{tabular}
\caption{Training settings. All rows but the last three are those of the released code of verl-agent, GiGPO and
GraphGPO, except 3,072 (release: 2,048) and GiGPO w/ std on WebShop (release: w/o std). $^{*}$In the codebase of GraphGPO.}
\label{tab:hyper}
\end{table}

\paragraph{Baselines.} GRPO, GiGPO and GraphGPO are run from their released code with the settings of their
releases, except as stated below. GiGPO's release divides its step-level term by the standard deviation on ALFWorld (w/ std) but not on
WebShop (w/o std); we use the w/ std variant on both. To our knowledge, the numbers reported in
the papers come from the trainer's validation (128 to 512 episodes, depending on the paper); on this validation,
our GraphGPO runs are at the level of the paper. Our rows are reproductions and need not equal the published numbers; for fairness, \ours{} is always compared with our own run of its base method. The HGPO row is that paper's setting with two history steps, as in all our runs (\cref{tab:hyper}).

\paragraph{The codebase of GraphGPO and its observation.} In \cref{tab:main}, GRPO, GiGPO and \ours{} added to them run in
the verl-agent codebase of Feng et al.~\cite{feng2025gigpo}; GraphGPO and \ours{}$_{\text{GraphGPO}}$ run in the release of
Cheng et al.~\cite{cheng2026graphgpo} (WebShop runs of the other methods inside that release: \cref{tab:ws}). That release
also renders a richer observation than verl-agent: in ALFWorld every observation ends with the agent's location, the object in its hand and the list of
objects it has moved; in WebShop the page is rendered with its buttons marked and the clicks made so far listed. We call the released functions of GraphGPO from our trainer (a state graph
per task group; step returns $10\,\gamma^{d}$ at distance $d$ from success, with $\gamma$ as in \cref{tab:hyper}, normalised
within groups of identical states; plus the episode advantage) and add the credit after its normalisation. The added
fields are removed before a page or observation of this release enters the log, so
the decision model sees the same text in both codebases.

\paragraph{Training games and seeds.} The eight environments of a task slot share a seed, so the
rollouts of a task play the same game, and two runs with the same seed on the same machine play the same games in
the same order. We therefore group runs into \emph{game sets} (A, B, C):
within a game set, all methods draw the same 16 of ALFWorld's 3,553 training games at every iteration. GRPO, GiGPO and \ours{}$_{\text{GRPO}}$
have one run on each of the three game sets (shown separately in \cref{fig:app_curves_sets}); \ours{}$_{\text{GiGPO}}$ and \ours{}$_{\text{GraphGPO}}$ also have three runs each, and GraphGPO has one. WebShop runs differ by seed (in \cref{tab:main}, three per method and two for GRPO; in \cref{tab:ws}, rows without $\pm$ are single runs).

\paragraph{Evaluation.} Besides the trainer's validation, we evaluate every final checkpoint outside the trainer on
episodes that are the same for every policy: 420 seen-room and 420 unseen-room ALFWorld episodes drawn from the
validation games, and 500 WebShop episodes. Each is played once (pass@1) with the trainer's
environment manager, prompt, parser, step limit and sampling temperature (0.4). The main text reports this evaluation; policies trained in the codebase of GraphGPO are evaluated with its
observation. Differences between methods are computed per episode, with intervals
from a bootstrap over games (\cref{app:runs}). WebShop numbers use the standard goal split of verl-agent; because most of
its evaluation goals also occur in training, we repeat the core comparison on a split whose test goals are
never trained on (\cref{app:ws}).

\paragraph{WebShop: train and test goals.} The verl-agent codebase uses WebShop's set of 1,000 products with 6,910 goals; it validates on ``goals 0--499'' and trains on the others, but it
shuffles the goal list with the seed of each environment, and its validation environments are seeded differently from
its training environments. The two index ranges are therefore not a split: 91\% of the validation goals of an
environment are training goals of a given training group. To our knowledge, the published numbers from the released code are obtained in this way,
and so are ours in \cref{tab:main}. For the held-out split we shuffle once for all environments, so that the 500
test goals form the same list everywhere and are never trained on, and we remove the 10 training goals whose
instruction also occurs among the test goals. Since a goal is a product with one combination of option values, test
and training goals still share products. Numbers from the two splits are not comparable with each other.

\clearpage
\subsection{The Request to the Decision Model}
\label{app:request}

A trajectory needs one or more requests: each holds at most 24 questions, so a fifty-step episode needs
three. The state is a JSON object with three fields: \texttt{task} (the goal sentence of the first observation),
\mbox{\texttt{first\_observation}} (at most 1,500 characters) and \texttt{log}, the list of steps. A step is the string
``\emph{action} $\rightarrow$ \emph{observation}'', where the action is the text the environment executed (the
content of the \texttt{<action>} tag, lower-cased; an output without the tag is shown as ``(no \texttt{<action>} in the output: invalid action)'') and
the observation is what the environment returned, cut at 400 characters. The observation of the last step is
replaced by ``(last step: the task is completed)'' or ``(last step: the episode ended, task not completed)''. In ALFWorld, question $k$ reads:

\begin{quote}\small
Step $k$ of the agent's log is: \emph{(the step)}. The agent acts in a household and has to complete the task in
the state. Considering the task and everything earlier in the log, did THIS step make progress toward completing
the task? Answer true for a useful new move: going to or opening a place that has not been searched yet and could
hold an object the task needs, picking up an object the task needs, bringing it to where the task needs it, or
using the right appliance on it. Answer false if nothing happens, if the step repeats an earlier one or returns to
a place that was already searched without a reason, if it handles an object the task does not need, or if it does
not bring the task closer to completion in any other way.
\end{quote}

The reworded question of the ablation (\cref{tab:abl}) replaces everything after ``\ldots\ the task in the state.'' with: ``The log is complete: you can see what
happened after this step. Did THIS step help the task? Answer true in three cases. (1) Exploration: the step went
to or opened a place that had not been searched yet and could hold an object the task needs. (2) Handling: the
step picked up an object the task needs, used the right appliance on it, or put it where the task asks. (3) A move
that was used: the step went to a place -- also one visited before -- and the agent then took, used or placed an
object the task needs there. Answer false if nothing happens, if the step repeats an earlier one, if it goes to a
place that was already searched and nothing the task needs is done there, if it handles an object the task does
not need, or if it only looks around.''

The answer to each question is a probability. The same text serves all six task types; the first wording was written once, checked on 240 trajectories that mix
expert and random actions (\cref{tab:offline}), and then kept.

\paragraph{WebShop.} The state and the request are built in the same way, with the instruction as the task. A page
is a list of fields. Cut at 400 characters, a results page would show only one or two of its ten products, and a product
page with many option values would not reach its title; for WebShop we therefore cut every field on its own (90
characters) and the page at 1,300, and rewrite a product page to begin with ``Product: \emph{title}
(\emph{price})''. The site returns the same page, without a message, for a click that has no target. The log therefore adds
``(there is no `\ldots' to click on this page: nothing happens)'' when the clicked text is not among the texts of
the page (a comparison with the environment's own list of clickable elements shows that every clickable element
appears as a text of the page),
``(the same click as before: nothing changes)'' when the same text is clicked again with nothing in between that
could have changed the selection, and ``(not an action of the site: nothing happens)'' for an output that the site
does not parse as a search or a click. The selected options are not shown in the text of a page, so a click on a
listed option value gets no note unless it repeats the click just before it: clicking grey, black, grey makes three
selections.
The notes follow the site's own parser (a greedy match of \emph{name}[\emph{argument}], ignoring text after the
bracket); they are how the decision model is told that a step had no effect. The last step carries ``(last step: the purchase matches the instruction)'' or ``(last step: the episode ended without a matching purchase)''. Question $k$ reads:

\begin{quote}\small
Step $k$ of the agent's log is: \emph{(the step)}. The agent shops on a website and has to buy a product that
matches the instruction in the state: the kind of item, its attributes, the options asked for (such as color and
size) and the price limit. Considering the instruction and the whole log, did THIS step bring the agent closer to
buying such a product? Answer true for a search whose query targets the item and its key attributes, clicking a
result that plausibly matches the instruction, selecting an option value the instruction asks for, going back or
to another page when the current results or product clearly do not match, or clicking buy now on a matching
product with the requested options. Answer false for a search unrelated to the instruction or repeating an earlier
one, clicking a product that does not match, selecting an option the instruction does not ask for, moving between
pages without a reason, an action that has no effect, or buying a product that does not match. The log is complete
and its last line says whether the episode ended with a matching purchase. Use it: if it did, the product that was
bought was the right one to open and its selected options were the right ones; if it did not, judge from the
instruction which step went wrong -- the product that was opened, an option that was selected or left out, or the
price. A search, a page turn or a return to the search page counts only if the agent then opened a product from
the page it led to; searching, turning pages or going back again and again without opening a product is false. In
the log a results page lists every product as its code, its title and its price, click[\emph{code}] opens that
product, and a product page starts with `Product:' followed by its title and price, then the option names and
values that can be clicked.
\end{quote}

The final wording adds the sentences on the outcome note and on browsing to a first wording, which four pilot
training runs used. It was chosen among four wordings on scripted episodes whose steps are labelled by
construction: for every kind of action, it separates right from wrong steps of that kind better than the first wording.

\clearpage
\subsection{Implementation Notes}
\label{app:impl}

\paragraph{The rule of the ablation.} The rule assigns 0 to a step whose observation is ``Nothing happens.'', whose output
has no valid action, or that repeats an earlier step of the episode verbatim (same action and same observation),
and 1 to every other step. Its scores are standardised like those of the decision model and given the same weight.

\paragraph{When the decision model is unreachable.} Failed requests to the decision model are retried; a run that still cannot score a
batch stops with an error instead of continuing without the credit, since a zero credit would be
the underlying method under another label.

\paragraph{Artefacts.} Our code is based on verl-agent (with the released code of GiGPO and GraphGPO), which builds
on verl and vLLM. These and Qwen2.5-1.5B-Instruct are distributed under Apache-2.0, and ALFWorld and WebShop under the MIT
licence; we use them for research, as intended. The decision model (Jev~1.13~\cite{typesafe2026jev}) is accessed
through its API under its terms of service. The four error-location benchmarks of \cref{sec:qual} are public
and used for evaluation only.

\clearpage
\subsection{Compute, Time and Cost}
\label{app:cost}
\begin{figure}[!htbp]
\centering
\includegraphics[width=\textwidth]{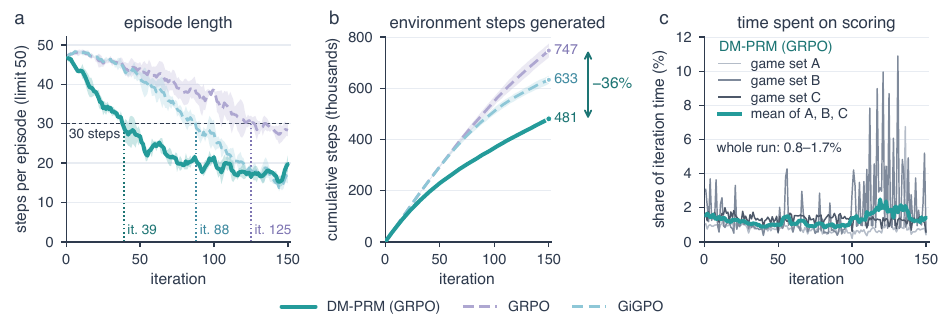}
\caption{Episodes shorten earlier with \ours{}$_{\text{GRPO}}$, so a run generates fewer environment steps; scoring takes
little time (ALFWorld; mean over three game sets, band: one standard deviation). (a) Episode length; dotted: first
iteration below 30 steps. (b) Cumulative environment steps; right: run totals (thousands) and difference to GRPO.
(c) Scoring share per iteration. (a) and mean in (c): 5-iteration moving average.}
\label{fig:app_dynamics}
\end{figure}

\begin{table}[!ht]
\centering\small
\setlength{\tabcolsep}{3.5pt}
\begin{tabular}{llrrrrr}
\toprule
 & & & env.\ steps & \multicolumn{2}{c}{scoring} & API \\
\cmidrule(lr){5-6}
Game set (GPU) & Method & hours & ($10^3$) & s / it. & \% time & (\$) \\
\midrule
A (RTX A6000) & GRPO & 69.9 & 778 & 0 & 0.0 & 0.00 \\
 & GiGPO & 61.7 & 627 & 0 & 0.0 & 0.00 \\
\rowcolor{oursbg} & \ours{}$_{\text{GRPO}}$ & 59.9 & 488 & 10 & 0.8 & 19.03 \\
\midrule
B (RTX A6000) & GRPO & 64.1 & 736 & 0 & 0.0 & 0.00 \\
 & GiGPO & 60.9 & 622 & 0 & 0.0 & 0.00 \\
\rowcolor{oursbg} & \ours{}$_{\text{GRPO}}$ & 59.2 & 481 & 13 & 1.7 & 18.75 \\
\midrule
C (RTX PRO 6000) & GRPO & 28.5 & 728 & 0 & 0.0 & 0.00 \\
 & GiGPO & 26.6 & 651 & 0 & 0.0 & 0.00 \\
\rowcolor{oursbg} & \ours{}$_{\text{GRPO}}$ & 24.6 & 473 & 7 & 1.3 & 18.46 \\
\bottomrule
\end{tabular}

\caption{Cost of a 150-iteration ALFWorld run on one GPU, from the trainer logs. \emph{hours}: wall-clock time of
training; \emph{env.\ steps}: environment steps generated; \emph{scoring}: median seconds per iteration spent on the
requests to the decision model and their share of the run's time; \emph{API}: cost of these requests. The run with
the credit needs no further GPU and is the shortest of the three on every game set.}
\label{tab:cost}
\end{table}

\paragraph{No extra compute.} \ours{} adds no model to the trainer: no value network is fitted and only the
policy is updated, so a run needs the same single GPU as the method it is added to. The only addition is the
scoring request to the decision model after the rollouts of an iteration. \Cref{tab:cost} gives the cost of the
150-iteration ALFWorld runs. Scoring an iteration takes about ten seconds, a small share of
a run's time, and costs about \$0.12 in API calls (\$0.039 per
thousand scored steps; about \$0.04 per WebShop iteration). Because episodes end
sooner once the policy improves, a run with the credit generates more than a third
fewer environment steps than the GRPO run on the same games and more than a fifth
fewer than the GiGPO run, and on every game set it finishes in less wall-clock time than either.

\paragraph{Where the saving comes from.} \Cref{fig:app_dynamics} follows the runs of \cref{tab:cost}
over training. On each game set the three methods start from the same episode length. \ours{}$_{\text{GRPO}}$
brings the smoothed mean length below 30 steps within the first third of training, GiGPO in the second half
and GRPO only in the last fifth. GiGPO catches up only near the end and GRPO not at all, so the saving in
steps accrues over most of the run, while scoring remains a small part of every iteration.

\paragraph{Budget of the study.} A WebShop run takes 12--21 hours on an NVIDIA RTX PRO 6000 (96 GB) and 28--47 hours
on an RTX A6000 (48 GB); ALFWorld runs are listed in \cref{tab:cost}. The runs behind \cref{tab:main} amount to about \GpuHours{}
GPU-hours of training, and to about \GpuHoursAll{} with pilots, ablations and variants.

\paragraph{Reproducing the runs.} Everything except the decision model is public: the policy model, the two benchmarks
and the trainer (verl-agent with the released code of GiGPO and GraphGPO). We will release our patches to the trainer, the
reward manager with the question texts of \cref{app:request}, the evaluation scripts, the lists of
evaluation episodes and the per-episode evaluation logs. The decision model is reached through its public API under the model name and version given in
\cref{sec:method}.

\clearpage
\subsection{Prompts of the Policy}
\label{app:prompts}

The policy is prompted with the unchanged templates of verl-agent (braces: fields filled in at every step; the
WebShop template has the same form).

\begin{quote}\footnotesize
\textbf{ALFWorld.} You are an expert agent operating in the ALFRED Embodied Environment. Your task is to:
\{task\_description\} Prior to this step, you have already taken \{step\_count\} step(s). Below are the most recent
\{history\_length\} observations and the corresponding actions you took: \{action\_history\} You are now at step
\{current\_step\} and your current observation is: \{current\_observation\} Your admissible actions of the current
situation are: [\{admissible\_actions\}]. Now it's your turn to take an action. You should first reason
step-by-step about the current situation. This reasoning process MUST be enclosed within \texttt{<think>}
\texttt{</think>} tags. Once you've finished your reasoning, you should choose an admissible action for current
step and present it within \texttt{<action>} \texttt{</action>} tags.
\end{quote}

\clearpage
\section{Additional Experiments}
\label{app:additional}

\subsection{Standard Deviations and Paired Comparisons}
\label{app:runs}

\begin{table}[!htbp]
\centering\small
\setlength{\tabcolsep}{3pt}
\resizebox{\textwidth}{!}{\begin{tabular}{lcccccccccc}
\toprule
 & \multicolumn{8}{c}{ALFWorld success (\%)} & \multicolumn{2}{c}{WebShop (\%)} \\
\cmidrule(lr){2-9}\cmidrule(lr){10-11}
Method & Pick & Look & Clean & Heat & Cool & Pick2 & All & Unseen & Score & Succ. \\
\midrule
{\color{cited}GPT-4o} & {\color{cited}75.3} & {\color{cited}60.8} & {\color{cited}31.2} & {\color{cited}56.7} & {\color{cited}21.6} & {\color{cited}49.8} & {\color{cited}48.0} & {\color{cited}46.0$^{a}$} & {\color{cited}31.8} & {\color{cited}23.7} \\
{\color{cited}Gemini-2.5-Pro} & {\color{cited}92.8} & {\color{cited}63.3} & {\color{cited}62.1} & {\color{cited}69.0} & {\color{cited}26.6} & {\color{cited}58.7} & {\color{cited}60.3} & {\color{cited}50.5$^{a}$} & {\color{cited}42.5} & {\color{cited}35.9} \\
\midrule
Qwen2.5-1.5B-Instruct & 13.4 & 5.9 & 6.7 & 0.0 & 5.6 & 8.3 & 7.6 & 4.8 & 22.5 & 4.4 \\
{\color{cited}ReAct} & {\color{cited}17.4} & {\color{cited}20.5} & {\color{cited}15.7} & {\color{cited}6.2} & {\color{cited}7.7} & {\color{cited}2.0} & {\color{cited}12.8} & {\color{cited}--} & {\color{cited}40.1} & {\color{cited}11.3} \\
{\color{cited}Reflexion} & {\color{cited}35.3} & {\color{cited}22.2} & {\color{cited}21.7} & {\color{cited}13.6} & {\color{cited}19.4} & {\color{cited}3.7} & {\color{cited}21.8} & {\color{cited}--} & {\color{cited}55.8} & {\color{cited}21.9} \\
{\color{cited}PPO} & {\color{cited}64.8$_{\pm 3.5}$} & {\color{cited}40.5$_{\pm 6.9}$} & {\color{cited}57.1$_{\pm 4.9}$} & {\color{cited}60.6$_{\pm 6.6}$} & {\color{cited}46.4$_{\pm 4.0}$} & {\color{cited}47.4$_{\pm 1.9}$} & {\color{cited}54.4$_{\pm 3.1}$} & {\color{cited}--} & {\color{cited}73.8$_{\pm 3.0}$} & {\color{cited}51.5$_{\pm 2.9}$} \\
{\color{cited}RLOO} & {\color{cited}88.3$_{\pm 3.0}$} & {\color{cited}52.8$_{\pm 8.6}$} & {\color{cited}71.0$_{\pm 5.9}$} & {\color{cited}62.8$_{\pm 8.7}$} & {\color{cited}66.4$_{\pm 5.5}$} & {\color{cited}56.9$_{\pm 4.7}$} & {\color{cited}69.7$_{\pm 2.5}$} & {\color{cited}68.7$_{\pm 10.7}$$^{a}$} & {\color{cited}73.9$_{\pm 5.6}$} & {\color{cited}52.1$_{\pm 6.7}$} \\
GRPO & 78.6$_{\pm 4.5}$ & 38.2$_{\pm 15.6}$ & 74.7$_{\pm 9.6}$ & 74.7$_{\pm 2.3}$ & 61.4$_{\pm 7.5}$ & 49.4$_{\pm 10.0}$ & 66.3$_{\pm 3.6}$ & 63.7$_{\pm 8.6}$ & 77.5$_{\pm 3.7}$ & 59.1$_{\pm 1.0}$ \\
GiGPO & 90.5$_{\pm 5.5}$ & 57.8$_{\pm 4.5}$ & 92.0$_{\pm 2.7}$ & 94.7$_{\pm 9.2}$ & 86.1$_{\pm 7.5}$ & 81.1$_{\pm 5.9}$ & 86.3$_{\pm 3.1}$ & 86.6$_{\pm 1.1}$ & 81.5$_{\pm 8.3}$ & 66.1$_{\pm 7.6}$ \\
{\color{cited}HGPO} & {\color{cited}--} & {\color{cited}--} & {\color{cited}--} & {\color{cited}--} & {\color{cited}--} & {\color{cited}--} & {\color{cited}92.8$_{\pm 1.1}$} & {\color{cited}90.2$_{\pm 0.8}$} & {\color{cited}85.6$_{\pm 2.9}$} & {\color{cited}71.5$_{\pm 4.0}$} \\
GraphGPO & 83.0 & 70.6 & 98.7 & 100.0 & 88.8 & 86.7 & 88.6 & 88.3 & 84.4$_{\pm 4.3}$ & 73.8$_{\pm 3.3}$ \\
\midrule
\rowcolor{oursbg}\ours{}$_{\text{GRPO}}$ & 97.6$_{\pm 1.4}$ & 91.2$_{\pm 5.1}$ & 95.1$_{\pm 5.0}$ & 94.0$_{\pm 4.0}$ & 89.1$_{\pm 4.3}$ & 66.7$_{\pm 11.5}$ & 90.0$_{\pm 3.4}$ & 92.9$_{\pm 1.4}$ & 84.6$_{\pm 2.6}$ & 72.2$_{\pm 0.7}$ \\
\rowcolor{oursbg}\ours{}$_{\text{GiGPO}}$ & 99.4$_{\pm 1.0}$ & 92.2$_{\pm 7.4}$ & 92.0$_{\pm 3.5}$ & 92.7$_{\pm 4.2}$ & 95.1$_{\pm 2.8}$ & 67.8$_{\pm 12.7}$ & 91.3$_{\pm 3.1}$ & 93.5$_{\pm 1.9}$ & 85.7$_{\pm 1.6}$ & 69.6$_{\pm 4.1}$ \\
\rowcolor{oursbg}\ours{}$_{\text{GraphGPO}}$ & 100.0$_{\pm 0.0}$ & 85.3$_{\pm 0.0}$ & 96.0$_{\pm 3.5}$ & 97.3$_{\pm 4.6}$ & 94.0$_{\pm 1.3}$ & 82.8$_{\pm 1.9}$ & 94.0$_{\pm 1.3}$ & 93.6$_{\pm 4.0}$ & 86.3$_{\pm 2.0}$ & 75.9$_{\pm 2.2}$ \\
\bottomrule
\end{tabular}
}
\caption{Extended version of \cref{tab:main}: the same rows with standard deviations, and two prompted agents (\%, pass@1). Black: our runs, mean $\pm$ sample standard deviation over runs (no $\pm$: a single run). {\color{cited}Grey: as reported}, with the $\pm$ of the paper where it gives one: GPT-4o~\cite{gpt4o}, Gemini-2.5-Pro~\cite{gemini25}, the policy model prompted with ReAct~\cite{react} and with Reflexion~\cite{Reflexion}, PPO and RLOO~\cite{rloo}, all from Feng et al.~\cite{feng2025gigpo}; HGPO (its setting with two steps of history, as in all our runs) and $^{a}$ from He et al.~\cite{he2026hgpo}. --: not reported there.}
\label{tab:runs}
\end{table}

\begin{figure}[!htbp]
\centering
\includegraphics[width=\textwidth]{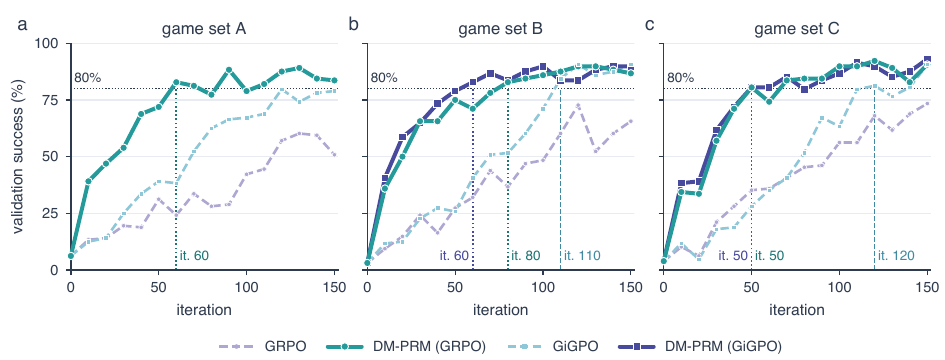}
\caption{The three ALFWorld runs separately (validation success, 128 episodes):
\ours{}$_{\text{GRPO}}$ is ahead of GRPO and GiGPO at every validation from iteration 10 through iteration 110 in
each of them. Vertical lines: first validation at or above 80\% (dashed: GiGPO). \ours{}$_{\text{GiGPO}}$: \AlfGjPairSets{}; its third run is on a further set.}
\label{fig:app_curves_sets}
\end{figure}

\Cref{tab:runs} extends \cref{tab:main} with standard deviations and with the policy model prompted with ReAct and with Reflexion.
\Cref{fig:app_curves_sets} shows the three ALFWorld runs separately. The gains are stable: on unseen rooms, the runs of
\ours{}$_{\text{GRPO}}$ differ far less from each other than those of GRPO do.

\paragraph{Paired differences.} Differences are computed per episode on the shared evaluation episodes, with 95\%
intervals from 10,000 resamples of games. \ours{}$_{\text{GRPO}}$ is
\AlfJevGrpoPairId{} points above GRPO on seen rooms (\AlfJevGrpoPairIdLo{} to \AlfJevGrpoPairIdHi{}) and
\AlfJevGrpoPairOod{} points above it on unseen rooms (\AlfJevGrpoPairOodLo{} to \AlfJevGrpoPairOodHi{}). Every other paired
difference is positive as well, on both benchmarks: \ours{}$_{\text{GRPO}}$ against GiGPO,
\ours{}$_{\text{GiGPO}}$ against the GiGPO runs on the same games (\AlfGjPairSets{}; the gain row of
\cref{tab:main} compares the means over all three runs of each method), and \ours{}$_{\text{GraphGPO}}$
against GraphGPO. The intervals lie above zero for the comparison with GraphGPO on seen and unseen rooms and, on unseen
rooms, for both comparisons with GiGPO.

\paragraph{Gains by task type.}\label{app:two} The gain over GRPO is positive on every task type, on seen and
unseen rooms (\cref{fig:alf}b,~c), and it is largest, about fifty points, where GRPO is weakest: examining
an object under a lamp on seen rooms, and that task and the two-object task on unseen rooms. On unseen rooms
\ours{}$_{\text{GRPO}}$ is also ahead of GiGPO on \AlfJevGigpoTypesAheadOod{} task types, most of all on the two-object
task, where GiGPO is weakest. The tasks that an outcome-only method solves least often are those on which it is
rewarded least often; a credit for every useful step exists even without that reward.

\clearpage
\subsection{WebShop}
\label{app:ws}

\begin{table}[!htbp]
\centering\footnotesize
\setlength{\tabcolsep}{3pt}
\resizebox{\textwidth}{!}{\begin{tabular}{lcccccc}
\toprule
 & GRPO & GiGPO & GraphGPO & \ours{}$_{\text{GRPO}}$ & \ours{}$_{\text{GiGPO}}$ & \ours{}$_{\text{GraphGPO}}$ \\
\midrule
won & 59.1 & 66.1 & 73.8 & 72.2 & 69.6 & 75.9 \\
goal's product, wrong options / price & 10.5 & \phantom{0}9.0 & \phantom{0}6.5 & \phantom{0}8.1 & 10.3 & \phantom{0}6.9 \\
other product; goal's was listed & 14.4 & 10.1 & \phantom{0}7.2 & \phantom{0}7.4 & 11.5 & \phantom{0}5.1 \\
other product; goal's not listed & 11.3 & \phantom{0}9.3 & \phantom{0}5.3 & \phantom{0}5.1 & \phantom{0}7.7 & \phantom{0}4.3 \\
no purchase & \phantom{0}4.7 & \phantom{0}5.5 & \phantom{0}7.3 & \phantom{0}7.2 & \phantom{0}0.9 & \phantom{0}7.9 \\
\bottomrule
\end{tabular}
}
\caption{WebShop: share of the 500 evaluation episodes of the final checkpoints (\%) by how they end, mean over
the runs of a method (the methods of \cref{tab:main}; single episodes in the codebase of GraphGPO: \cref{app:qualws}).}
\label{tab:wsloss}
\end{table}

\begin{table}[!htbp]
\centering\small
\setlength{\tabcolsep}{5pt}
\begin{tabular}{lcc}
\toprule
Method & Score & Succ. \\
\midrule
\multicolumn{3}{l}{\emph{Standard goal split}} \\
GRPO & 77.5$_{\pm 3.7}$ & 59.1$_{\pm 1.0}$ \\
GiGPO & 81.5$_{\pm 8.3}$ & 66.1$_{\pm 7.6}$ \\
GraphGPO & 84.4$_{\pm 4.3}$ & 73.8$_{\pm 3.3}$ \\
\rowcolor{oursbg}\ours{}$_{\text{GRPO}}$ & 84.6$_{\pm 2.6}$ & 72.2$_{\pm 0.7}$ \\
\rowcolor{oursbg}\ours{}$_{\text{GiGPO}}$ & 85.7$_{\pm 1.6}$ & 69.6$_{\pm 4.1}$ \\
\rowcolor{oursbg}\ours{}$_{\text{GraphGPO}}$ & 86.3$_{\pm 2.0}$ & 75.9$_{\pm 2.2}$ \\
\midrule
\multicolumn{3}{l}{\emph{Test goals never used in training}} \\
GRPO & 81.7 & 69.2 \\
\rowcolor{oursbg}\ours{}$_{\text{GRPO}}$ & 84.4 & 77.0 \\
\midrule
\multicolumn{3}{l}{\emph{Inside the codebase of GraphGPO}} \\
GRPO & 76.7 & 65.8 \\
GiGPO & 87.0 & 70.6 \\
\rowcolor{oursbg}\ours{}$_{\text{GRPO}}$ & 85.9$_{\pm 2.9}$ & 74.9$_{\pm 2.9}$ \\
\rowcolor{oursbg}\ours{}$_{\text{GiGPO}}$ & 87.6 & 74.8 \\
\bottomrule
\end{tabular}

\caption{WebShop, final checkpoints: task score (the benchmark's reward for the purchased product) and success
(a task score of 1), \%, mean $\pm$ sample standard deviation over runs (no $\pm$: a single run).}
\label{tab:ws}
\end{table}

\begin{figure}[!ht]
\centering
\includegraphics[width=\textwidth]{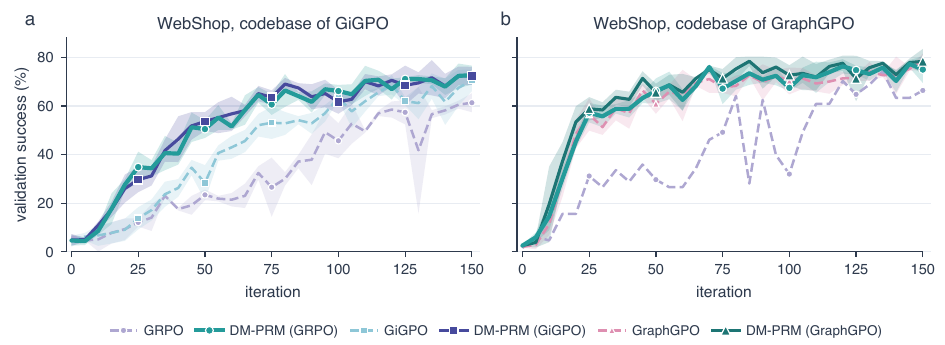}
\caption{WebShop, validation success during training (128 episodes; band: one standard
deviation over runs). (a) Codebase of GiGPO, which also runs GRPO: \ours{}$_{\text{GRPO}}$ and \ours{}$_{\text{GiGPO}}$ learn faster than GiGPO. (b) Codebase of GraphGPO, with GRPO run in it as well: \ours{}$_{\text{GRPO}}$ is above GRPO from iteration 10 on.}
\label{fig:app_ws_curves}
\end{figure}

\begin{figure}[!ht]
\centering
\includegraphics[width=\textwidth]{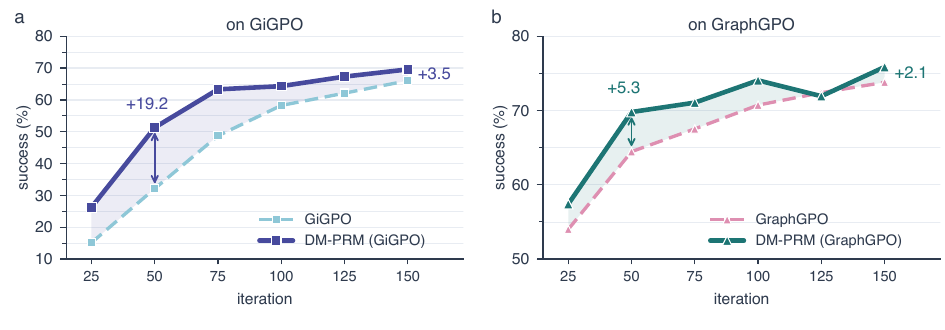}
\caption{WebShop, success of the checkpoints on the 500 evaluation episodes (means over runs): GiGPO and GraphGPO
without (dashed) and with \ours{} (solid), as in \cref{fig:ws}a (each panel with its own y-axis). Numbers: lead of \ours{} at iteration 50 and at
the end of training.}
\label{fig:app_ws_ckpt}
\end{figure}

\paragraph{Faster learning.} \ours{}$_{\text{GRPO}}$ is ahead of GRPO on held-out goals at every evaluated
checkpoint, by about twenty points at iterations 25 and 50. On the standard split it passes 50\% validation success
about 25 iterations before GiGPO (\cref{fig:app_ws_curves}a), as does \ours{}$_{\text{GiGPO}}$, which is ahead
of the GiGPO it is added to at nearly every validation; in the codebase of GraphGPO the validation success of
\ours{}$_{\text{GRPO}}$ is above that of GRPO from iteration 10 on (\cref{fig:app_ws_curves}b). On the 500
evaluation episodes, the credit also leads when added to the other two methods (\cref{fig:app_ws_ckpt}):\WsAppCkptClause{}
At the end of training, \ours{}$_{\text{GRPO}}$ is above GRPO on held-out goals and, inside the codebase of
GraphGPO (three runs), well above GRPO and above GraphGPO itself (\cref{tab:ws}).

\paragraph{Baseline and weight of the credit.} \Cref{tab:kinds} gives the mean score by kind of action in the
first iteration, which is the check of \cref{sec:method}. In ALFWorld the kinds that complete a task score above the
mean of all steps. In WebShop purchases score low and searches high, so the pooled baseline of
\cref{eq:group} teaches the agent to buy less (in a pilot run, purchases fell from 39 to 23 of 128 episodes in three iterations, but not with a baseline per kind), whereas under \cref{eq:kind} every kind of action has a mean
credit of zero. At $w=1$ the credit dominates the update and the agent learns to buy late, so we use $w=0.3$.
Both choices were made on short pilot runs on held-out goals, before the runs of \cref{tab:main}. On ALFWorld, $w=0.3$ learns more slowly in the first 20 iterations (22.7\% against 35.9\% validation success at iteration 10) and is ahead at iteration 60 (\cref{tab:abl}).

\clearpage
\subsection{The Decision Model as a Judge of Steps}
\label{app:judge}

\paragraph{Cost against an LLM judge.}\label{app:bench} \Cref{tab:bench} compares the cost of scoring with the
decision model and with a frontier LLM prompted as a judge, on the same sampled items of the four error-location
benchmarks of \cref{sec:qual}. An item is one task with all its steps: the judge names the earliest wrong step in one request, and the decision model scores every step. The LLM judge is tens to hundreds of times more expensive and tens of times
slower, and it returns text that has to be parsed. The decision model returns one probability per step, at a cost and a delay that are small next to those of training (\cref{app:cost}), which is
what a scorer inside a training loop needs.

\begin{table}[!ht]
\centering\small
\setlength{\tabcolsep}{5pt}
\begin{tabular}{lcc}
\toprule
 & LLM judge & Decision model \\
\midrule
\$ per 1,000 items & 10.7--27.2 & 0.06--0.73 \\
median latency (s) & 7.3--10.3 & 0.25--0.32 \\
\bottomrule
\end{tabular}
\caption{Cost of scoring with the decision model and with an LLM judge (Claude Sonnet~5, Anthropic; temperature 0, 4,000 output
tokens) on the same sampled items of four error-location benchmarks (300 per benchmark, 180 for Who\&When; ranges over the benchmarks; costs as billed in October 2026).}
\label{tab:bench}
\end{table}

\paragraph{The scores judge the step, not the episode.}\label{app:val} The decision model is told how the episode ended, so its
scores could echo the outcome. They do not: over the rollouts of the first 25 iterations of an ALFWorld run, the
kind of step explains several times more of the variance of the step scores than the outcome of the episode does
(39\% against 7\%), almost all of the variance of the credit lies within episodes, and an action without effect
receives the same low score in won and lost episodes. Nor does the policy learn to please the decision model without
solving tasks: in these iterations, a lost episode almost never has a higher mean step score than the median won
episode.

\clearpage
\section{Additional Qualitative Analysis}
\label{app:qual}

\subsection{ALFWorld: What the Decision Model Says About Single Steps}
\label{app:cases}

\begin{figure}[!htbp]
\centering
\includegraphics[width=\textwidth]{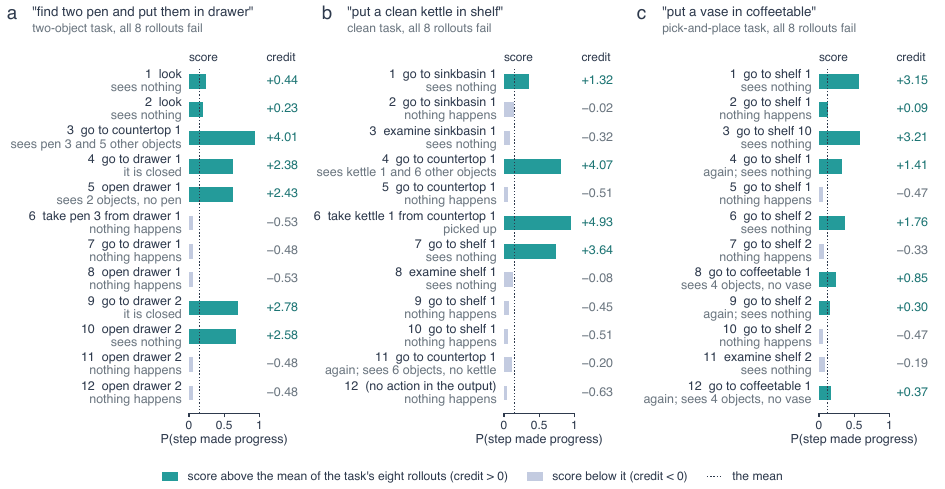}
\caption{The decision model tells the steps of failed rollouts apart where the outcome cannot: three rollouts
of the untrained policy from ALFWorld task groups in which all eight rollouts fail (first 12 of 50 steps). Bars: the
decision model's probability that the step made progress; right: the credit. Grey: the observation in short form
(\emph{again}: the place was visited before). For each task type with such a group, other than the type of \cref{fig:case}a,
we show the rollout whose 12 shown steps have the largest gap in mean score between steps with and without an effect.}
\label{fig:qual_scorer}
\end{figure}

\Cref{fig:qual_scorer} shows three rollouts of the untrained policy from groups in which all
eight rollouts fail. The scores follow what a step does: arriving where the pen or the kettle lies and taking the kettle all score
above 0.8, and the steps answered with ``Nothing happens'' score below 0.15; in panel~c no vase is
found, yet every return to a place scores lower than the first visit. Over the nine such groups of iteration 1
(3,600 steps), steps that go somewhere new, open or handle an object make up 21\% of all steps but 93\% of the tenth with the
highest credit, and in 70 of the 72 rollouts the highest-scored step is an action with an effect. \Cref{tab:qual_scorer} gives the mean score by what a step does, before training and
later in training: the kinds that serve a task stay well above returns, looking and steps without effect. \Cref{tab:offline} gives the scores of selected kinds of steps before any training, on trajectories that mix the
environment's expert actions with random actions.

\begin{table}[!ht]
\centering\footnotesize
\setlength{\tabcolsep}{4pt}
\begin{tabular}{lrrrr}
\toprule
 & \multicolumn{2}{c}{iteration 1} & \multicolumn{2}{c}{iteration 17} \\
\cmidrule(lr){2-3}\cmidrule(lr){4-5}
what the step does & steps & score & steps & score \\
\midrule
go to a new place & 569 & 0.48 & 1,396 & 0.45 \\
heat / cool / clean / lamp & 115 & 0.38 & 29 & 0.47 \\
put & 126 & 0.37 & 469 & 0.33 \\
take & 190 & 0.37 & 519 & 0.37 \\
open / close & 232 & 0.31 & 657 & 0.43 \\
go to a place again & 998 & 0.14 & 2,157 & 0.14 \\
look / examine & 1,359 & 0.09 & 6 & 0.07 \\
no effect / invalid & 2,230 & 0.06 & 348 & 0.06 \\
\midrule
all steps & 5,819 & 0.15 & 5,581 & 0.28 \\
\bottomrule
\end{tabular}
\caption{Mean score of the decision model by what a step does, with the number of steps, over the 128 training
rollouts of iteration 1 (untrained policy) and of iteration 17 of one ALFWorld run.}
\label{tab:qual_scorer}
\end{table}

\begin{table}[!ht]
\centering\footnotesize
\setlength{\tabcolsep}{3pt}
\begin{tabular}{lrr}
\toprule
Kind of step & score & $n$ \\
\midrule
take an object the task names & 0.916 & 217 \\
take another object & 0.074 & 86 \\
clean / cool / heat, right appliance & 0.93--0.96 & 63 \\
last step of a won episode & 0.922 & 108 \\
go to a place, first visit & 0.436 & 2,364 \\
go to a place, revisit & 0.170 & 1,204 \\
look / inventory / examine & 0.104 & 860 \\
\bottomrule
\end{tabular}
\caption{Scores of selected kinds of steps on mixed expert and random ALFWorld trajectories (5,814 steps), before any
training.}
\label{tab:offline}
\end{table}

\clearpage
\subsection{ALFWorld: What the Agents Do}
\label{app:qualalf}

\begin{figure}[!htbp]
\centering
\includegraphics[width=\textwidth]{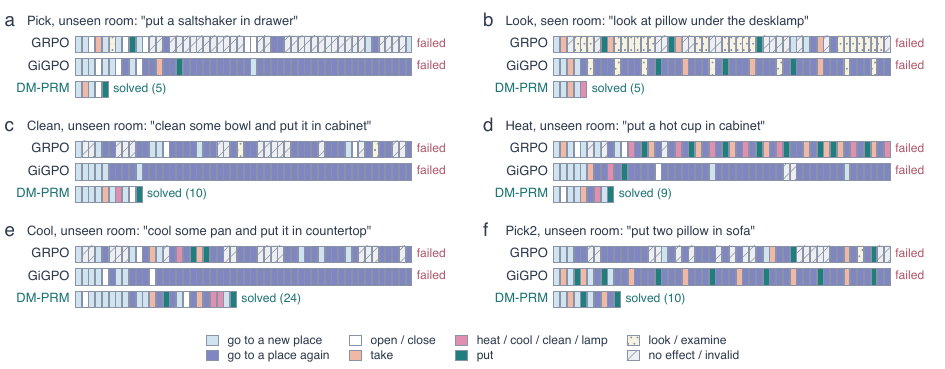}
\caption{Six episodes that only \ours{}$_{\text{GRPO}}$ solves, one per task type (final policies of one run per method on the
same game, one tile per step; a failed episode uses all 50 steps): its policy moves on to new places, while the
strips of the baselines fill with returns to places already visited (GiGPO) and with returns and actions without
effect (GRPO). For each task type, we show the shortest episode that our policy solves and both baselines fail; (d) is another episode of the task of \cref{fig:case}b (same room, objects placed differently).}
\label{fig:qual_alf}
\end{figure}

\Cref{fig:qual_alf} extends the case study of \cref{fig:case} to every task
type. In these six episodes, 30 of our policy's 38 \emph{go to} steps lead to a new place. GiGPO's policy walks among three to
seven places, and GRPO's spends about a third of its steps on actions without effect. Neither baseline reaches the place
where ours finds the object (a, c, e) or the lamp (b); in (d) both heat a mug, not a cup, and in (f) each handles
only one of the two pillows. These six are short by selection, but the pattern holds beyond them and for all three RL methods: over all evaluation
episodes, the policies trained with the credit go to new places more often than they return, whereas the baselines do the opposite
(\cref{tab:beh}); within solved episodes, too, they return less and go to new places more than any baseline. More than three times as many episodes
are solved only by \ours{}$_{\text{GRPO}}$ as only by GiGPO (\cref{tab:qual_alf_excl}). Against GiGPO, the gain lies in how many episodes are
solved: \ours{}$_{\text{GRPO}}$ solves more within 30 and within 50 steps (\cref{tab:qual_alf_eff}).

\begin{table}[!ht]
\centering\small
\setlength{\tabcolsep}{3.5pt}
\begin{tabular}{lrrrrrrr}
\toprule
 & & & \multicolumn{3}{c}{all episodes} & \multicolumn{2}{c}{solved} \\
\cmidrule(lr){4-6}\cmidrule(lr){7-8}
 & episodes & steps & return & new & no effect & return & new \\
\midrule
GRPO & 2{,}520 & 27.2 & 42.8 & 14.0 & 17.0 & 28.9 & 22.3 \\
GiGPO & 2{,}520 & 15.7 & 47.6 & 23.7 & 2.2 & 27.8 & 32.9 \\
GraphGPO & 840 & 13.4 & 48.0 & 26.8 & 1.3 & 22.6 & 39.1 \\
\midrule
\ours{}$_{\text{GRPO}}$ & 2{,}520 & 15.7 & 24.2 & 37.5 & 3.0 & 17.4 & 41.5 \\
\ours{}$_{\text{GiGPO}}$ & 2{,}520 & 13.2 & 26.8 & 35.3 & 0.6 & 17.2 & 41.1 \\
\ours{}$_{\text{GraphGPO}}$ & 2{,}520 & 11.9 & 27.7 & 37.6 & 0.4 & 16.1 & 42.9 \\
\bottomrule
\end{tabular}

\caption{ALFWorld: how the final policies spend their steps on the evaluation episodes (runs of \cref{tab:main}).
\emph{steps}: mean per episode; shares of steps (\%) that return to a place visited earlier in the episode, go to a
new place, or have no effect; the last two columns are restricted to solved episodes.}
\label{tab:beh}
\end{table}

\begin{table}[!ht]
\centering
\begin{minipage}[t]{\dimexpr(\textwidth-1.5em)/2\relax}
\centering\small
\setlength{\tabcolsep}{6pt}
\begin{tabular}[t]{lrr}
\toprule
Solved by & seen & unseen \\
\midrule
only GRPO & 6 & 0 \\
only GiGPO & 35 & 26 \\
only \ours{}$_{\text{GRPO}}$ & 92 & 103 \\
\midrule
all but GRPO & 248 & 287 \\
all but GiGPO & 25 & 25 \\
all but \ours{}$_{\text{GRPO}}$ & 36 & 22 \\
\midrule
all three & 769 & 756 \\
none & 49 & 41 \\
\midrule
episodes & 1{,}260 & 1{,}260 \\
\bottomrule
\end{tabular}
\caption{ALFWorld evaluation episodes by which of the three final policies solve them (three runs; the policies
play the same episodes). \emph{All but} X: solved by the other two and not by X.}
\label{tab:qual_alf_excl}
\end{minipage}\hfill
\begin{minipage}[t]{\dimexpr(\textwidth-1.5em)/2\relax}
\centering\small
\setlength{\tabcolsep}{3pt}
\begin{tabular}[t]{@{}lrrrrrr@{}}
\toprule
 & \multicolumn{4}{c}{solved within (\%)} & \multicolumn{2}{c}{mean steps} \\
\cmidrule(lr){2-5}\cmidrule(lr){6-7}
 & 10 & 20 & 30 & 50 & all & solved \\
\midrule
\multicolumn{7}{l}{\emph{seen rooms}} \\
GRPO & 35.4 & 51.6 & 58.8 & 66.3 & 26.3 & 14.2 \\
GiGPO & 62.9 & 79.4 & 83.3 & 86.3 & 15.3 & 9.8 \\
\ours{}$_{\text{GRPO}}$ & 53.4 & 77.6 & 85.2 & 90.0 & 15.7 & 11.9 \\
\midrule
\multicolumn{7}{l}{\emph{unseen rooms}} \\
GRPO & 27.9 & 47.5 & 54.8 & 63.7 & 28.1 & 15.7 \\
GiGPO & 56.4 & 78.6 & 82.3 & 86.6 & 16.1 & 10.9 \\
\ours{}$_{\text{GRPO}}$ & 47.2 & 77.9 & 86.4 & 92.9 & 15.8 & 13.2 \\
\bottomrule
\end{tabular}
\caption{Share of the ALFWorld evaluation episodes (\%; final policies, three runs) solved within 10, 20, 30 and 50
steps (50 is the limit), and mean steps per episode over all and over solved episodes.}
\label{tab:qual_alf_eff}
\end{minipage}
\end{table}

\clearpage
\subsection{WebShop: What the Agents Do}
\label{app:qualws}

\begin{figure}[!htbp]
\centering
\includegraphics[width=\textwidth]{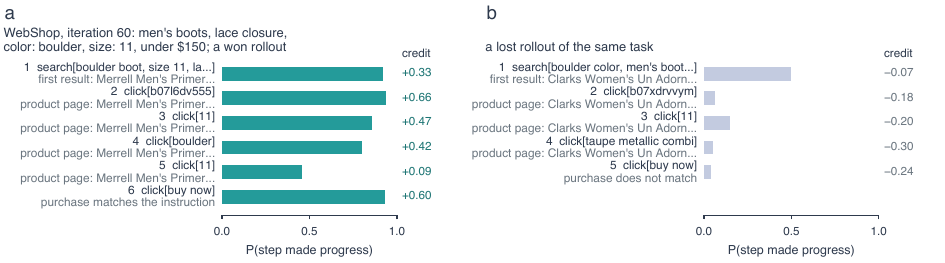}
\caption{WebShop: (a) a won and (b) a lost rollout of the same task (iteration 60 of training). Bars:
probability that the step made progress; right: the credit times the weight ($w\,c_{i,k}$, $w=0.3$).}
\label{fig:scores}
\end{figure}

\paragraph{Scores of single steps.} \Cref{fig:scores} shows a won and a lost rollout of the same task, which differ from the
first search on. The opened product and the purchase score above 0.9 in the won rollout and below 0.1 in the
lost one. Because the credit compares a step with the steps of its own kind (\cref{eq:kind}), the lost
purchase is set against other purchases, not against searches. In full, the kinds of \cref{eq:kind} are: a search on the search page; on a results page, opening a product, turning the page (next or previous: one kind) and going back to the search page; on a product page, an option click, opening a sub-page, buying, and leaving by ``$<$ prev'' or by ``back to search'' (two kinds); on a sub-page, these two ways of leaving. A step without effect counts as a search on the search page, as opening a product on a results page, as an option click on a product page and as ``$<$ prev'' on a sub-page.

\begin{figure}[!htbp]
\centering
\includegraphics[width=\textwidth]{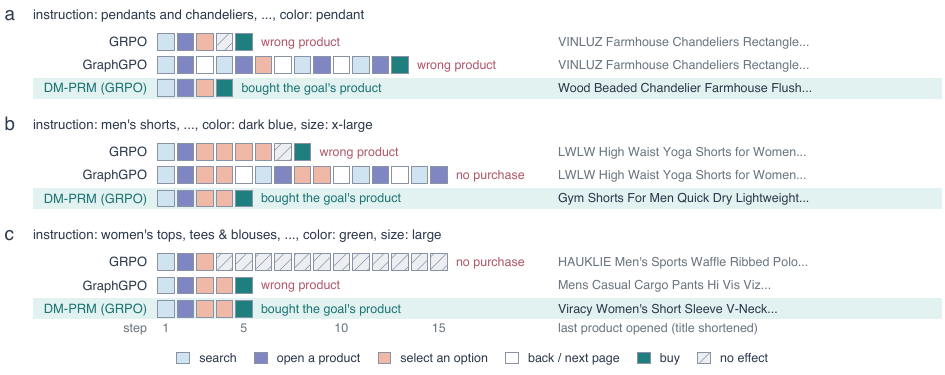}
\caption{WebShop, final policies (codebase of GraphGPO, the first run of each method) on the same episodes: where the
other two buy another product or do not buy, \ours{}$_{\text{GRPO}}$ opens the goal's product and buys it
in at most five steps. One tile per step, coloured by what the step does; after each strip: how the episode ends.
Right: the last product opened. Shown: three episodes that our policy wins and the other two lose.}
\label{fig:qual_ws}
\end{figure}

\paragraph{Final policies.} \Cref{fig:qual_ws} shows three episodes. In (a) the searches of GRPO
and \ours{}$_{\text{GRPO}}$ both list the goal's product first. Ours opens it, selects \emph{pendant} and
buys it; GRPO buys the second entry, a chandelier above the price limit, as GraphGPO does. In (b)
the other two policies open the women's yoga shorts at the top of their lists; our policy's shorter query lists the goal's men's
shorts second, and it opens them. In (c) GRPO opens a men's polo shirt and repeats one option click until the
step limit, and GraphGPO buys men's cargo pants; our policy buys the requested top.

\end{document}